\documentclass[11pt]{article}
\usepackage[]{acl}
\usepackage{times}
\usepackage{latexsym}
\usepackage[T1]{fontenc}
\usepackage[utf8]{inputenc}
\usepackage{microtype}
\usepackage{inconsolata}
\usepackage{graphicx}
\usepackage{tabularx}
\usepackage{booktabs}
\usepackage{footmisc}
\usepackage{hyperref}
\usepackage{xspace}
\usepackage{makecell}
\usepackage{siunitx}
\usepackage{subcaption}  
\usepackage{multirow}
\usepackage{rotating}
\usepackage{annotates}
\providecommentcommand{rom}{blue}{Roman}
\providecommentcommand{chr}{green!30!black}{Christopher}
\providecommentcommand{car}{violet}{Carina}

\newcommand{\multemo}{Mult$^2$EMo\xspace}
\newcommand{\emotion}[1]{\textsl{#1}\xspace}

\title{Emotion Experience, Expression, and Perception: \\ Emotion Analysis on Multimodal Social Media Posts}

\author{Christopher Bagdon$^{1}$, Carina Silberer$^{2}$, \and Roman Klinger$^{1}$\\
  $^{1}$Fundamentals of Natural Language Processing, University of Bamberg, Germany\\
  $^{2}$Institut f\"ur Maschinelle Sprachverarbeitung, University of Stuttgart, Germany\\
  \texttt{\{christopher.bagdon,roman.klinger\}@uni-bamberg.de}\\
  \texttt{CarinaSilberer@posteo.de}}

\begin{document}
\maketitle
\begin{abstract}
  Emotions are an essential aspect of human communication, particularly on social media, where authors frequently combine text and images to convey their emotions. Yet prior work on emotion analysis of social media posts has overlooked two important aspects in regard to measuring how well readers can reconstruct the authors' intent: (1)~the image modality, with most work focusing solely on text, and (2)~the real-world events that trigger the expressed emotions, and their relationship to the post content. We therefore study the relation between (a) the author's experience of the event that caused them to write a social media post and (b) the content of the post, with a focus on readers' capability to 
  reconstruct that emotion expression. To do that,  we introduce the Multimodal Multi-Emotion-Model dataset (\multemo), created by collecting annotations from both authors and readers on the posts and their triggering events. We find that reconstruction is possible but challenging for both human readers and computational models. We show that understanding the triggering event is crucial for accurate reconstruction, and that reconstruction is particularly challenging when posts rely heavily on the image to express emotion.
\end{abstract}

\section{Introduction}
Personal communication via a combination of text and images is a
recent phenomenon, which gained popularity with the rise of the
Internet and further spread with the advent of social media and
smartphones. In recent years, multimodal communication on social media
has become increasingly prevalent
\citep{illendula2019multimodal,image_stats}.  As such, understanding
how authors express emotions in multimodal posts is essential for
understanding modern emotion communication and how people use social
media.

In natural language processing (NLP), emotion analysis involves
interpreting emotions expressed in text. The emotions are authors'
private states, and the texts are authors' attempts to share their
private states with others
\cite{wilson-wiebe-2005-annotating}. Gaining access to these private
states for corpus creation can be done in several ways: by asking
authors to label their own emotions
\cite{kajiwara-etal-2021-wrime,bagdon-etal-2025-donate}, inferring
them from information in the text such as hashtags
\cite{mohammadEmotionalTweets2012,mohammad-bravo-marquez-2017-emotion},
or by asking third-party readers to reconstruct them
\cite{demszky-etal-2020-goemotions,liu-etal-2013-joint}, with the
latter being most common in NLP.
\begin{figure}
    \centering
    \includegraphics[width=\linewidth]{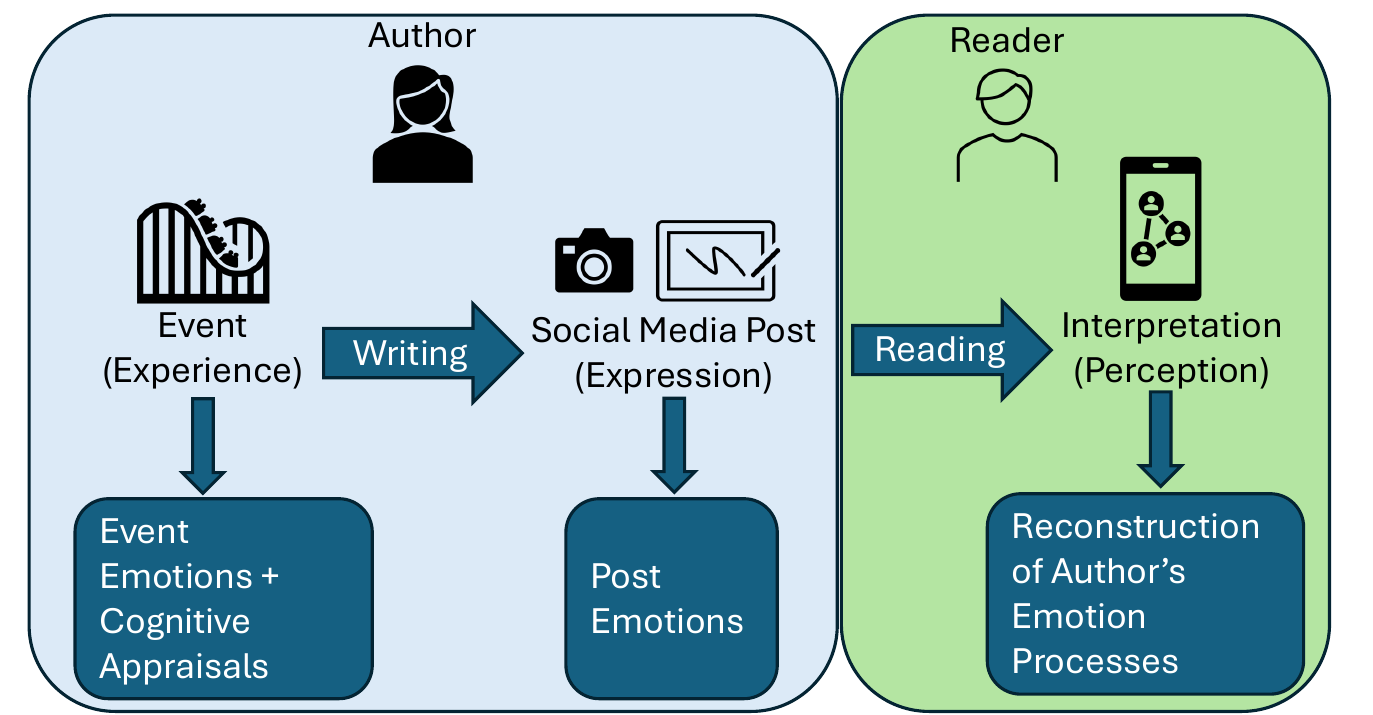}
    \caption{The process of emotion communication on social
      media. Authors experience an event which triggers an emotion,
      and they express that emotion in a multimodal post. Readers then
      interpret the author's expression of emotion.}
    \label{fig:emotion_diagram}
\end{figure}
Such reconstruction of the author's emotion comes with inaccuracies,
due to differing interpretations of the text -- an aspect that has
been studied already \citep{Troiano2023}. It is, however, unclear if
these inaccuracies stem only from the reader's varying interpretation
of the author's intended emotion, or if authors might also be
inaccurate or make assumptions regarding prior knowledge (e.g.,~about the inciting event that led to an experienced emotion) that leads to
a challenging interpretation of their posts. 

We therefore study three aspects, shown in Figure~\ref{fig:emotion_diagram} -- experience, expression and perception  -- jointly, and do so in a multimodal setup: 
How do the \textit{experienced} and the \textit{expressed}  emotion differ
(referred to as \textit{emotion processes} in this paper)? How does
that difference affect readers' \textit{perception} of the author's emotion processes? We investigate this by focusing on the
author's cognitive evaluation of the event and hence the emotion
experience \cite{scherer, klinger-2023-event}, and by comparing authors' descriptions of inciting events to readers' interpretations of the same events. 

More concretely, we answer the following research questions: (RQ1)~Is a
multimodal post sufficient to reconstruct authors' intended expression
of emotion and their appraisal of an event? (RQ2)~How is emotion
reconstruction impacted by the relatedness of the event and the post?
(RQ3)~How does image type and relation to the text impact emotion reconstruction?

Our study introduces two fundamental novelties: (a) we not only
study the social media post content, but also the events that cause
the author to write them, and (b) we study multimodal social media
posts in breadth, while previous work typically focused on narrower 
domains such as memes
\citep{sharma-etal-2020-semeval,meme_emotion_detection_sharma2024,Shi2025EnhancingME}.

To support these investigations, we introduce the Multimodal Multi-Emotion-Model dataset (\multemo), comprising 7,200 social media posts. Authors annotate each post for the emotions experienced in response to a triggering event and their cognitive appraisals, a description of that event, and the emotions expressed in the post. Additionally, readers annotate a subset of 1,440 posts, attempting to reconstruct all author annotations. 

We find that readers and models can reconstruct authors' expression of
emotion and appraisal of an event to some extent, though models
perform better than readers, especially on reconstructing the
cognitive evaluation of an event, measured through appraisal
variables. For an accurate reconstruction, a crucial element is an aligned interpretation of the triggering event.  Both humans and models
rely more on the textual information than on the image modality.

\section{Related Work}
\subsection{Emotion Theories in NLP}
Emotion theories are important aspects of emotion analysis in NLP, as
they provide frameworks for understanding emotion processes. For
example, in emotion classification, texts are labeled either by using
categorical
frameworks 
or dimensional frameworks such as Valence--Arousal--Dominance
\citep{russell1980circumplex}. Categorical approaches can use
coarse-grained taxonomies such as Ekman's (\citeyear{ekman1999basic})
six basic emotions \cite{mohammadEmotionalTweets2012} or more
fine-grained taxonomies such as \citet{demszky-etal-2020-goemotions}
who use 27~emotion categories. These annotations can be single labels
or multi-labels \cite{bostan-klinger-2018-analysis}, and can also
include intensity ratings \citep{mohammad-bravo-marquez-2017-emotion}.

However, these frameworks do not capture the full complexity of
emotion processes, as they do not account for the cognitive evaluation
of events, or appraisals, which are an important aspect of emotion
processes \citep{scherer}. Appraisals are subjective evaluations of
events based on personal values, motivation, and context, and play a
crucial role in shaping emotional responses. \citet{scherer}
formalizes them in the \textit{Component Process Model}, which
conceptualizes emotion as a dynamic process in which an event triggers
a cascade of appraisal checks -- evaluations of, for instance, the
event's novelty, its goal relevance, and one's ability to cope with it
-- whose outcomes jointly produce the emotion experience. Crucially,
because appraisals are subjective, the same event can elicit different
emotions in different individuals.

\subsection{Emotion Experience, Expression, and Perception}
Emotion research in NLP can focus on various aspects: 
the emotion experienced by the author, 
the emotion expressed by the author, 
the reader's 
perception of the author's emotion processes,
and the emotion experienced by the reader in response to the author's expression.  
Studies often cover two aspects simultaneously, however it is rare for more than two to be studied together. \multemo contains annotations for the first three aspects.

\paragraph{Experience and Expression.} Emotion experience is a private state elicited by an event, while emotion expression is the attempt to share that private state, which can be conveyed through modalities such as text and images \cite{wilson-wiebe-2005-annotating}. Both are important for understanding emotion communication, as authors may express emotions differently than they experience them.

Emotion experience is studied using both categorical and dimensional frameworks. Studies such as \citet{scherer1997isear} and \citet{Troiano2023} ask participants to recall and describe an event in which they felt a target emotion (e.g., ``I felt joy when\dots'') and answer appraisal questions about the event. \citet{bagdon-etal-2025-donate} ask participants to label both the emotion they experienced in response to an event and the emotion they later expressed in a social media post about the event. \citet{yeo-jaidka-2025-beyond} use appraisal information and emotion labels to test 
large language models on \textit{emotion reasoning}, finding that they are poor at associating event outcomes with specific emotions.

Emotion expression is the most common aspect of emotion research in NLP; the majority of emotion labeled datasets attempt to capture the emotion expressed by the author in a text \cite{bostan-klinger-2018-analysis}. This can be done by asking authors to label the emotion they expressed \cite{kajiwara-etal-2021-wrime,bagdon-etal-2025-donate,li-etal-2025-third} or via distant labeling methods
 \citep{mohammadEmotionalTweets2012,mohammad-bravo-marquez-2017-emotion,liu-etal-2013-joint}. Annotations directly from authors are more accurate and can capture instances in which emotion is implicitly conveyed, however they are more time-consuming and expensive to collect.

\paragraph{Perception.} 
Readers' perception of the author's emotion expression is a common approach to emotion classification in NLP \cite{strapparava-mihalcea-2007-semeval,demszky-etal-2020-goemotions,klinger-2023-event,liu-etal-2013-joint,Troiano2023,buechel-hahn-2017-emobank
}, as this annotation process is easily accessible and can be done at scale via crowdsourcing. Previous studies ask third-party readers to label texts for the emotion expressed by the author. This has been used as a stand-in for author annotations, however, recent work shows that readers are not proficient at reconstructing authors' emotion processes \citep{Troiano2023,li-etal-2025-third}. \citet{li-etal-2025-third} found that third parties' ability to reconstruct authors' private states is limited when using both fine and coarse-grained emotion taxonomies; however, readers who belonged to the author's social group performed better. 

	\begin{table*}[ht!]
	\centering
	\small
	\begin{tabular}{lcccccccccccc}
		\toprule
		& \multicolumn{3}{c}{Reader $\uparrow$} & \multicolumn{3}{c}{CLIP $\uparrow$} & \multicolumn{3}{c}{CLIP Rand. $\uparrow$} &\multicolumn{3}{c}{Qwen3 $\uparrow$}\\
		\cmidrule(r){2-4} \cmidrule(){5-7} \cmidrule(l){8-10} \cmidrule(l){11-13}
		& RR & R & V & T & I & T+I & T & I & T+I & T & I & T+I \\
		\cmidrule(r){2-4} \cmidrule(){5-7} \cmidrule(l){8-10} \cmidrule(l){11-13}
		Anger & .52 & .44 & .47 & .43 & .27 & .45 & .46 & .09 & .46 & .54 & .36 & .52\\
		Disgust & .44 & .38 & .40 & .39 & .23 & .36 & .40 & .08 & .38 & .29 & .20 & .35\\
		Fear & .52 & .44 & .45 & .53 & .32 & .56 & .55 & .17 & .53 & .58 & .33 & .59\\
		Joy & .82 & .58 & .58 & .58 & .41 & .58 & .59 & .15 & .58 & .54 & .42 & .56\\
		Sadness & .67 & .60 & .63 & .60 & .31 & .60 & .58 & .02 & .58 & .57 & .33 & .61\\
		Surprise & .44 & .35 & .34 & .48 & .25 & .50 & .48 & .20 & .49 & .30 & .14 & .32\\
		\cmidrule(r){2-4} \cmidrule(){5-7} \cmidrule(l){8-10} \cmidrule(l){11-13}
		Macro Avg. & .57 & .47 & .48 & .50 & .30 & .51 & .51 & .12 & .50 & .47 & .30 &.49\\
		\bottomrule
	\end{tabular}
	\caption{F1 scores for emotion classification by readers and models. Reader F1 is measured by comparing individual readers (R) or majority vote (V) against the author, and across reader--reader pairs (RR). Model F1 reports mean scores from 3 training runs for baseline CLIP and CLIP with random images (CLIP Rand.), across text (T), image (I), and text + image (T+I).}
	\label{table:emotion_classification_simple}
\end{table*}

\subsection{Multimodal Emotion Analysis}

Multimodal emotion analysis, particularly combining text and images, remains underexplored relative to text-only work. Most multimodal work focuses on video, audio, and text combinations \cite{shou2025multimodallargelanguagemodels}. Of the work on image and text, many focus on memes \cite{sharma-etal-2020-semeval,meme_emotion_detection_sharma2024,Shi2025EnhancingME}, or sentiment analysis \cite{al2024comprehensive}. While these studies provide valuable insights into multimodal architectures and text--image relations, they do not address the interplay between emotion experience, expression, and perception that underlies emotion communication.

Emotion analysis faces multimodal modeling challenges such as modality collapse \cite{shou2025multimodallargelanguagemodels}, which is when a model relies too heavily on one modality (e.g., text) and ignores the other (e.g., image). 
Various studies on related tasks 
have proposed methods to address this issue.
For example, \citet{yang-etal-2024-uncertainty-guided} use uncertainty to rebalance modalities for multimodal hate speech detection in  posts, and \citet{WU2025111485} dynamically reweigh modalities and use multiple fusion stages for fake news detection.
However, these methods are not specifically designed for emotion analysis, and have not been applied here.
 
\section{Data Collection Methods}
\label{sec:datacollection}

We collect \multemo in two stages: (1)~We first collect posts directly from authors (Author Phase), such that we can capture the emotion experience, expression, and further context to best understand the author's intent. (2)~Then we collect reader annotations on posts collected during the Author Phase to understand the readers' perception of the author's expression and experience (Reader Phase).%
\footnote{The anonymized dataset and surveys are available upon request via \url{https://www.uni-bamberg.de/en/nlproc/projects/item}.}

\subsection{Author Phase}
\label{ssec:authorphase}
We recruit participants via Prolific\footnote{\url{https://www.prolific.com}} in multiple annotation tasks by emotion: anger, disgust, fear, joy, sadness, and surprise. A task consists of providing and annotating three social media posts. 
To select each post, we prompt participants to recall an event which both triggered the target emotion and that they wrote a multimodal social media post about.

After providing the post, participants share details about it in the following steps:\footnote{We share more details in Appendix~\ref{App:Dataset_collection}.} (1)~\textit{Event Details.} 
The event which triggered the emotion, 
emotion labels and intensities for the event, and 
how confident they are in recalling it. (2)~\textit{Appraisal}. Participants rate 21~appraisal dimensions for the event. (3)~\textit{Post Details.} 
Emotion labels and intensities for the post along with emotion stimulus information for the text. (4)~\textit{Image Details.} 
Description of the content of the image, label for the image type, and 
emotion stimulus information for the image. (5)~\textit{Text--Image Relation.} 
The rated relationship between the text and image in terms of how much they rely on each other for understanding and how much they each convey the target emotion. (6)~\textit{Final Post Questions.} Indication of  
the context 
needed to understand the post, 
the reason for posting, and the 
intended audience.

\subsection{Reader Phase}
We present potential participants annotation tasks by social media platform: Instagram, Facebook, Twitter (X), and ``Various Platforms''. Each task consists of annotating three posts collected in the author phase. 

As we want to understand how well readers interpret authors' emotions, the context, and intentions behind the posts, we ask participants to reconstruct authors' answers. For example, instead of rating how intensely they felt an
emotion, they rate how intensely they think the
author felt the emotion. 

\subsection{Data Statistics}
\multemo contains 7,200 multimodal social media posts, with 1,200
posts for each of the six emotions, annotated by 1,101 unique
authors. Participants share posts from a variety of platforms: Facebook
(47\%), Instagram (30\%), Twitter (X) (18\%), and ``Various Platforms'' 
(5\%). 
We collect reader annotations for 1,440 posts, 240 per emotion, balanced by platform. 
Each post is annotated by 3 readers, resulting
in 4,320 reader annotations. This subset is used as a test set for our experiments, while the remaining 5,760 posts are used as a training set.  

\section{Experiments}
We use \multemo to investigate readers' and models' ability to reconstruct authors' emotion processes (RQ1). Based on their performance, we analyze   event--post relatedness (RQ2) as well as the impact of images (RQ3).

\subsection{RQ1: Is a multimodal post sufficient to reconstruct an author's expression of emotion and their appraisal of an event?}
\label{sec:RQ1}

We address RQ1 via human readers (Section~\ref{sec:rq1_readers}) and model predictions (Section~\ref{sec:rq1_models}).

\subsubsection{Readers}
\label{sec:rq1_readers}
\paragraph{Experimental Setup.} 
We evaluate reader against author annotations using F1 for post emotion and root mean square error (RMSE) for appraisals. For emotion, we compare individual readers~(R), and majority vote~(V)\footnote{Ties are broken using readers' confidence scores.}, 
a common approach for 3rd party annotators, against the author, as well as  reader-to-reader~(RR), discussed in Section~\ref{sec:rq2}. For appraisals, we compute individual RMSE (each reader vs.\ the author) and mean RMSE (mean of three readers vs.\ author).\footnote{A majority vote approach for appraisals yielded too many unresolvable ties.} We compare against two baselines: random (scores sampled uniformly from 1--5) and mean (per-appraisal mean score in \multemo).

\paragraph{Results.}
Column~R in Table~\ref{table:emotion_classification_simple} shows how 
individual 
readers can reconstruct authors' emotion processes, 
while column~V shows how reader majority vote performs. Readers are best at reconstructing \emotion{joy} and \emotion{sadness}, and struggle to reconstruct \emotion{disgust} and \emotion{surprise}. The majority vote performs only slightly better than individual readers. 

We report the mean RMSE 
 of all appraisal dimensions in Figure~\ref{fig:appraisal_overall}.
Readers perform better than random, but comparably to the mean baseline, showing that readers are not proficient at reconstructing authors' appraisals. This observation differs across appraisal variables.

Readers perform better at reconstructing some appraisals, such as \emotion{pleasantness} and \emotion{goal support}, than others; however, there is no appraisal dimension in which individual readers outperform the mean of three readers, showing that individual readers are especially poor at reconstructing appraisals.\footnote{\label{foot:appraisal_overall}Results for each appraisal dimension in Appendix~\ref{sec:appraisal_appendix}.} Readers' overall poor performance is likely because appraisals are subjective and based on personal values and motivations, making them difficult for readers to reconstruct without access to the author's internal state or additional context regarding the author \cite{Troiano2023}.  

\subsubsection{Models}
\label{sec:rq1_models}
\paragraph{Experimental Setup.}
We separately train baseline models for emotion and appraisal prediction. All models are fine-tuned CLIP models \cite{radford2021learning}.\footnote{\label{footnote:modeling_details}Model details and hyperparameters in Appendix~\ref{app:modeldetails}.} For each task, we train models using three modalities: text-only~(T), image-only~(I), and text and image~(T+I). 
As a control condition, we rerun the models with the images randomly shuffled to 
test if the models exploit useful information from the images. We report average results across three runs with different random seeds, using default model parameters.
\begin{figure}[t]
    \centering
    \includegraphics[width=\linewidth]{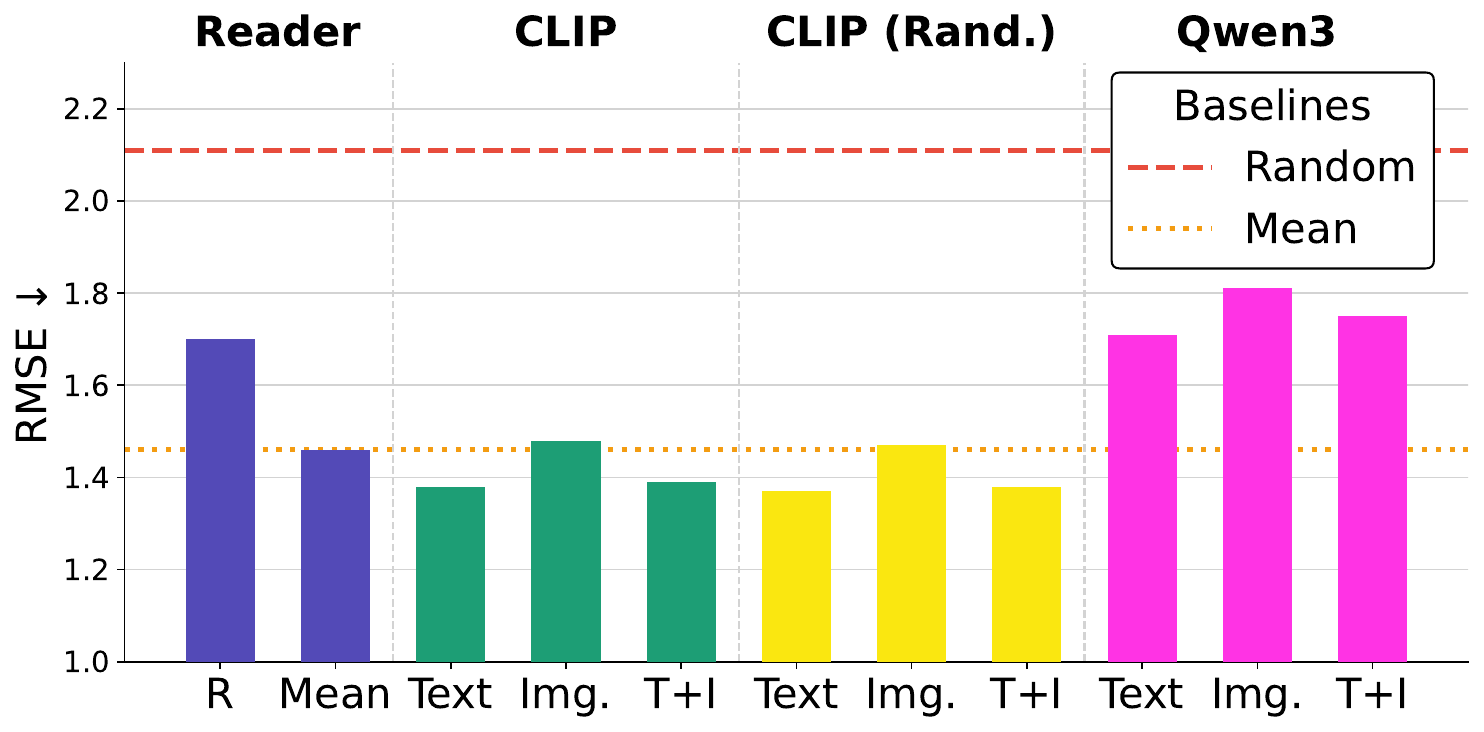}
    \caption{Mean RMSE for all appraisal dimension predictions. Baselines shown as dashed (Random) and dotted (Mean) lines. Reader R is individual readers.}
    \label{fig:appraisal_overall}
\end{figure}
\paragraph{Results.}
For emotion classification, we report macro-average F1 scores 
in Table~\ref{table:emotion_classification_simple}. For text and T+I models, \emotion{joy} and \emotion{sadness} are best predicted, while \emotion{disgust} and \emotion{anger} show lowest performance. For all emotions, the text-only (.5) and multimodal models (.51) perform considerably better than the image-only models (.3), however, the image-only models still perform above random chance, showing that 
images do contain emotion information. 

We do, however, not observe a performance difference between using images in addition to text irrespective of them being the correct or a random image (columns T+I), which indicates that the two modalities are generally not effectively combined. 

Figure~\ref{fig:appraisal_overall} shows the average RMSE results for appraisals, using the same random and mean baselines introduced in Section~\ref{sec:rq1_readers}.\footref{foot:appraisal_overall} Overall, all models perform better than chance, and the text-only and multimodal models perform better than the mean baseline, but the image-only models perform similarly to the mean baseline.  This shows that models are able to reconstruct some aspects of appraisals from text and images. \textit{Pleasantness} and \textit{goal support} perform best with RMSE scores below 1.2, despite their mean baseline RMSE being above 1.5, while other appraisals, such as \textit{others responsibility}, are closer to the mean baseline; in summary, the models are able to reconstruct some appraisals based on the information in the post.

Dimensions applicable to a wide range of events, such as \textit{pleasantness}, are easier to reconstruct than ambiguous ones like \textit{suddenness} (e.g., how sudden is reading a news article about a bombing?). This ambiguity likely affects both annotation consistency and model learning, as different interpretations of the appraisal question might lead to different scores on similar instances.

The random image models perform similarly to the regular image models, suggesting that the image modality is not effectively used for appraisal prediction. This is reinforced by the fact that the image-only models perform similarly with the regular and random image settings. However, it is unclear if this is because the image-only models are unable to exploit appraisal-relevant information from the images, or because the images do not contain much information about appraisals.

\subsubsection{Multimodal LLM}
To further investigate whether models can exploit information in images, we use a multimodal LLM (MLLM) to predict emotions and appraisals, as MLLMs pair the language understanding of a strong LLM backbone with visual comprehension acquired from pretraining on large amounts of image–text data.

\paragraph{Experimental Setup.}
We use a zero-shot approach with Qwen3-VL-8B-Instruct \cite{qwen3technicalreport}, run locally. 
We prompt for emotion classification and appraisal prediction separately, though we predict all appraisals together using a single prompt.\footnote{Prompts found in Appendix~\ref{app:modeldetails}.} To match our previous modeling experiment, we prompt every post three times and report the mean F1 scores from the 3 runs. To mitigate order bias in prompts, we randomize the list of emotions and list of appraisals within the prompts. 

\paragraph{Results.}
For emotion classification, the MLLM achieves macro-average F1 scores comparable to the multimodal CLIP model for all modalities, however it struggles with different emotions. In particular, it performs better on \emotion{anger} and worse on \emotion{disgust} and \emotion{surprise}. For appraisal prediction, the MLLM performs worse than both the CLIP model and the random image CLIP model. 

The comparable performance of the MLLM and the multimodal CLIP model on emotion classification suggests that the multimodal models rely on primarily the textual modality, with images contributing less to the overall performance, rather than the problem stemming from 
an ineffective integration of visual information.
The results are less supportive for appraisal prediction, where the MLLM performs worse than both the CLIP model and the random image CLIP model for all modalities. Previous research has also found that LLMs are not proficient at annotating ratings for subjective tasks \cite{bagdon-etal-2024-expert}.

\subsubsection{Reader vs. Model Comparison}
To gauge readers' and models' ability to reconstruct authors' emotion processes, we compare their performance on emotion classification and appraisal regression. 
 We compare reader majority vote~(V) and multimodal CLIP~(T+I) on emotion classification in Table~\ref{table:emotion_classification_simple}. Readers and models perform closely, with mean F1 of .48 and .51, respectively, and both find \emotion{joy} and \emotion{sadness} easiest to classify. Models outperform readers on \emotion{surprise} (.50 vs.\ .34) and \emotion{fear} (.56 vs.\ .45), while readers lead on \emotion{sadness} (.63 vs.\ .60) and \emotion{disgust} (.40 vs.\ .36).

For appraisal regression we compare reader and CLIP model results in Figure~\ref{fig:appraisal_overall}.\footref{foot:appraisal_overall} Overall, models perform better than human readers on every appraisal dimension, except for \emotion{pleasantness} which is very close (1.19 vs.~1.17 RMSE). The largest difference in performance is for \emotion{familiarity}, with models having an RMSE of~1.38 compared to~1.62 for readers. The differences are even more pronounced when comparing individual readers to models. 

We include reader majority vote and mean appraisal scores to assess how well readers and models can serve as proxies for author annotations, as these approaches are often used in practice \cite{bostan-klinger-2018-analysis}. For emotion classification, majority vote shows little improvement over individual readers, supporting previous findings that reader annotations are not reliable substitutes for author labels \cite{Troiano2023, li-etal-2025-third}. 

Furthermore, we do a statistical analysis based on \citet{hitchhikers}'s approach to significance testing for NLP. As F1 is a set-level metric with no per-instance decomposition, and each reader annotated only a small number of posts (4,320 annotations from 959 readers over 1,440 posts), per-annotator testing is statistically underpowered. We therefore treat the pooled reader annotations as a single system, evaluate readers and the model over the identical annotation instances (assigning the model its prediction for the corresponding post), and assess significance with a paired cluster bootstrap: posts are resampled with replacement (10,000 resamples), with all annotations of a sampled post kept together to account for the dependence between annotations of the same post. We compare the best performing multimodal model to readers.

Table~\ref{tab:model-vs-readers} shows that the model significantly outperforms individual readers. A per-post majority vote over readers narrows this gap and does not differ significantly from the model. These results are consistent with our broader finding that individual readers' perception of authors' emotions is noisy, while aggregation across readers recovers some of the divergence.

\begin{table}[t]
\centering
\small
\begin{tabular}{lcccc}
\toprule
 & $F_1$ & $\Delta F_1$ & 95\% CI & $p$ \\
\midrule
Model                & .51 & ---   & ---            & ---   \\
R   & .46 & .049  & $[.025, .075]$ & $< 10^{-4}$ \\
V       & .48 & .03   & $[-.01, .05]$  & .22   \\
\bottomrule
\end{tabular}
\caption{Statistical significance of emotion prediction model performance against individual readers (R) and a per-post majority vote over readers (V). $\Delta F_1$, confidence intervals, and $p$-values are relative to the model (two-sided).}
\label{tab:model-vs-readers}
\end{table}
For appraisals, reader mean RMSE improves substantially over individuals, yet closely mirrors the mean baseline -- as does image-only model performance. We suspect that both are mathematical artifacts: the reader mean flattens individual scores toward the centre of the 1--5 Likert scale, while image-only models, unable to exploit visual information, fall back on training data patterns, producing a similar effect.

Overall, we find that both readers and models are able to reconstruct authors' emotion processes to some extent, though it varies by emotion and appraisal. Models outperform 
readers, especially on appraisals, though for some 
dimensions neither perform much better than the mean baseline.

\subsection{RQ2: How is emotion reconstruction impacted by the relatedness of the event and the post?}
\label{sec:rq2}
As seen above, readers and models struggle to reconstruct authors' emotion processes -- but why? Social media posts are written after the triggering event, and readers are yet further removed, encountering it solely through the author's account. Information may thus be lost at two points: (1)~when writing, as authors may omit details about their experience, and (2)~when reading, as readers may misinterpret the expressed emotion or the event itself. To investigate this, we compare authors' experienced and expressed emotions (Section~\ref{sec:rq2_expressed}) and examine the impact of event understanding on reader (Section~\ref{sec:rq2_event}) and model performance (Section~\ref{Sec:Event Desc. Models}), using individual reader annotations and multimodal models, respectively.\footnote{Appendix~\ref{App: Event vs Post} further analyzes how post content relates to performance.}
\begin{table}[t]
  \small
	\centering
\begin{tabular}{lrrr}
\toprule
 & \multicolumn{3}{c}{\textbf{Mismatch}}   \\
 \cmidrule(lr){2-4} 
 Expressed Emotion& \%Posts & $\Delta$Reader & $\Delta$Model  \\
\cmidrule(l){2-4}
Anger & 13 & $-$.21 & $-$.06  \\
Disgust & 15 & $-$.12 & $-$.16 \\
Fear & 9 & $-$.31 & $-$.18  \\
Joy & 2 & $-$.46 & $-$.45  \\
Sadness & 7 & $-$.36 & $-$.31  \\
Surprise & 14 & $-$.01 & $-$.08 \\
\cmidrule(l){2-4}
Macro Avg. & 10 & $-$.25 & $-$.21 \\
\bottomrule
\end{tabular}
	\caption{
  Mismatch: posts where primary event and post emotions differ. $\Delta$R / $\Delta$M: difference in reader / model F1 between the full test set and the subset.}
	\label{tab:post_event_pct_comparison}
\end{table}

\subsubsection{Experienced vs. Expressed Emotion}
\label{sec:rq2_expressed}
The experience of an emotion during an event may differ from the expression of that emotion in a post about the event. We analyze how frequently this occurs and if performance is worse when it does.
\paragraph{Experimental Setup.}
We examine posts where the primary expressed and experienced emotions differ (Mismatch), comparing reader ($\Delta$R) and model ($\Delta$M) F1 on this subset against the full test set. Further analysis of emotion presence differences between event and post is in Appendix~\ref{App: Event vs Post}.
\begin{table}[t]
  \centering
\small
\begin{tabular}{clrrrr}
\toprule
& & \textbf{Q1} & \textbf{Q2} & \textbf{Q3} & \textbf{Q4} \\
\midrule
    & Mean Similarity & .02 & .21 & .39 & .59 \\
    & Std. & .02 & .06 & .05 & .08 \\
    & Min. & .01 & .10 & .30 & .48 \\
    & Max & .10 & .30 & .48 & .97 \\
\midrule
    \multirow{7}{*}{\rotatebox[origin=c]{90}{Post Emotion (F1)~$\uparrow$}}
    & Anger & .36 & .42 & .45 & .56 \\
    & Disgust & .23 & .38 & .37 & .51 \\
    & Fear & .36 & .38 & .54 & .63 \\
    & Joy & .41 & .56 & .65 & .71 \\
    & Sadness & .41 & .54 & .65 & .79 \\
    & Surprise & .27 & .38 & .34 & .42 \\
\cmidrule(l){3-6}
    & Macro Avg. & .34 & .44 & .51 & .61 \\
\midrule
    \multirow{12}{*}{\rotatebox[origin=c]{90}{Appraisals (RMSE)~$\downarrow$}}
    & Pleasantness & 1.72 & 1.38 & 1.14 & 0.99 \\
    & Unpleasantness & 1.82 & 1.53 & 1.35 & 1.18 \\
    & Not considered & 1.80 & 1.61 & 1.58 & 1.45 \\
    & Own responsibility & 1.70 & 1.53 & 1.44 & 1.29 \\
    & Own control & 1.67 & 1.48 & 1.45 & 1.39 \\
    & Goal relevance & 1.82 & 1.69 & 1.69 & 1.67 \\
    & Goal support & 1.68 & 1.51 & 1.46 & 1.35 \\
    & Event predictability & 1.71 & 1.71 & 1.67 & 1.56 \\
    & External standards & 1.74 & 1.59 & 1.54 & 1.44 \\
    & Internal standards & 1.78 & 1.62 & 1.44 & 1.46 \\
    & Effort & 1.83 & 1.67 & 1.68 & 1.58 \\
\cmidrule(l){3-6}
    & Macro Avg. & 1.81 & 1.70 & 1.65 & 1.60 \\
\bottomrule
\end{tabular}
\caption{Reader F1 and RMSE scores split by quartiles of author and reader event description similarity. Appraisal results only include dimensions with at least a 0.1 difference in RMSE between Q1 and Q4. Full results found in Appendix~\ref{Appendix:event_sim}.}
\label{table:post_event_similarity_bins}
\end{table}

\paragraph{Results.}
Table~\ref{tab:post_event_pct_comparison} shows that authors' primary experienced and expressed emotions often differ, and both readers and models struggle as a result, with F1 differing by .25 and .21 on mismatched posts, respectively. The difference is largest for \emotion{joy}, though sample sizes are small. Notably, models perform worse on mismatched \emotion{surprise} posts (\mbox{$-$.08 F1}) while readers show almost no difference, showing readers' difficulties with \emotion{surprise} are not mismatch-driven. Appendix~\ref{App: Event vs Post} further shows that authors tend to omit negative experienced emotions but rarely add new ones. 
Together, these results show that it is the primary emotion mismatch, rather than omissions, that drives performance drops, possibly reflecting authors' tendencies to express more socially acceptable emotions than they experience \cite{Hess_Hareli_2016}.

\subsubsection{Readers' Understanding of the Event}
\label{sec:rq2_event}
Post text may directly describe the triggering event, be tangentially related, or omit it entirely -- potentially leading readers to misinterpret the event. We assess whether this contributes to the difficulty of reconstructing authors' expressed emotions.
\paragraph{Experimental Setup.}

 \multemo's event description annotations allow us to compare authors' and readers' event descriptions, giving insight into how well readers infer the event from the post (full comparison in Appendix~\ref{App: Experienced vs Expressed emotion}). 
 We measure the similarity between author--reader event descriptions 
 using a cross-encoder~\cite{reimers-2019-sentence-bert} trained on the STS Benchmark~\cite{enevoldsen2025mmtebmassivemultilingualtext,muennighoff2022mteb}, then split posts into quartiles~(Q1--Q4) by 
 similarity to examine the relation between similarity and task performance.\footnote{Example posts with author and reader event descriptions and similarity scores in Figure~\ref{fig:example_posts}, Appendix~\ref{App:Dataset_collection}.}

 We bin by quartile rather than fixed similarity thresholds for two reasons. First, per-post F1 does not exist: F1 is computed over a set of instances, so relating similarity to classification performance requires partitioning the data. Second, cross-encoder similarity scores are ordinal rather than calibrated to an interpretable absolute scale; quantile binning depends only on the rank order of similarities and yields equally sized strata, ensuring per-quartile (and per-emotion) F1 estimates of comparable reliability.

\paragraph{Results.}

Table~\ref{table:post_event_similarity_bins} shows that higher reader--author event description similarity (i.e.,~the more accurately readers understand the event) correlates with better emotion classification and appraisal regression performance, with F1 scores at least 50\% higher in Q4 than Q1 (over twice as high for \emotion{disgust}). Furthermore, readers' event descriptions are more similar to each other than to authors' (see Appendix~\ref{App: Experienced vs Expressed emotion}), pointing to a consistent but often divergent understanding, and mirroring the reader--author vs.\ reader--reader agreement gap in Table~\ref{table:emotion_classification_simple}. Together, these results demonstrate that event understanding is a key factor in emotion reconstruction, and that information loss or omission during writing makes the task more difficult for readers.

\subsubsection{Post Text vs. Event Description (Models)}
\label{Sec:Event Desc. Models}
We saw that when readers better understand the event, they more accurately reconstruct emotions and appraisals, but what about models? 
To evaluate their understanding of events, we examine how using event descriptions instead of post texts affects model performance.

\paragraph{Experimental Setup.}
We train models replacing post text with author event descriptions, keeping all other settings identical.\footnote{Target emotion words are removed from event descriptions due to their inclusion in the annotation prompt.} We report mean F1 and RMSE across three runs for emotion classification and appraisal regression, respectively.\footref{footnote:modeling_details}

\begin{table}[t]
	\centering
  \small
\begin{tabular}{lcccccc}
\toprule
 & \multicolumn{3}{c}{Post Text $\uparrow$} & \multicolumn{3}{c}{Event Descrip. $\uparrow$} \\
 \cmidrule(r){2-4} \cmidrule(l){5-7}
Emotion & T & I & T+I & T & I & T+I \\
 \cmidrule(r){2-4} \cmidrule(l){5-7}
Anger & .43 & .27 & .45 & .48 & .26 & .43 \\
Disgust & .39 & .23 & .36 & .49 & .22 & .50 \\
Fear & .53 & .32 & .56 & .62 & .29 & .63 \\
Joy & .58 & .41 & .58 & .74 & .40 & .76 \\
Sadness & .60 & .31 & .60 & .62 & .26 & .62 \\
Surprise & .48 & .25 & .50 & .64 & .22 & .66 \\
 \cmidrule(r){2-4} \cmidrule(l){5-7}
Macro Avg. & .50 & .30 & .51 & .60 & .27 & .60 \\
\bottomrule
\end{tabular}
\caption{Emotion classification F1 on \multemo test set using post text or event descriptions as input, across text (T), image (I), and text + image (T+I) modalities.}
	\label{table:emotion_classification_event_description} 
\end{table}

{\setlength{\tabcolsep}{4pt}
	\begin{table}[t]
		\centering
		\small
		\begin{tabular}{lcccccc}
			\toprule
			& \multicolumn{3}{c}{Post Text $\downarrow$} & \multicolumn{3}{c}{Event Descrip. $\downarrow$} \\
			\cmidrule(lr){2-4} \cmidrule(lr){5-7}
			& T & I & T+I & T & I & T+I  \\
			\cmidrule(lr){2-4} \cmidrule(lr){5-7}
			Goal Support & 1.21 & 1.40 & 1.23 & 1.15 & 1.41 & 1.18 \\
			Internal standards & 1.29 & 1.48 & 1.28 & 1.24 & 1.50 & 1.24 \\
			Own control & 1.20 & 1.27 & 1.21 & 1.14 & 1.29 & 1.16 \\
			Pleasantness & 1.19 & 1.47 & 1.19 & 1.03 & 1.47 & 1.03 \\
			Unpleasantness & 1.26 & 1.48 & 1.26 & 1.17 & 1.50 & 1.18 \\
			\cmidrule(lr){2-4} \cmidrule(lr){5-7}
			Macro Avg. & 1.35 & 1.42 & 1.36 & 1.31 & 1.46 & 1.34 \\
			\bottomrule
		\end{tabular}
		\caption{Model performance (RMSE) for appraisal regression on Mult2Emo test set using event descriptions as input. Models are fine-tuned on text only (T), image only (I), and text + image (T+I). Only appraisal dimensions which differed by more than 0.05 RMSE are reported; full results found in Appendix~\ref{sec:model_eval_appendix}.}
		\label{table:appraisal_regression_event_description_short}
	\end{table}
}

\paragraph{Results.}
Tables~\ref{table:emotion_classification_event_description} and~\ref{table:appraisal_regression_event_description_short} show emotion classification and appraisal regression results.\footnote{Full results in Appendix~\ref{Appendix:event_sim}.} Models using event descriptions achieve higher emotion classification F1 scores than those using post text (excluding image-only models), with multimodal F1 at .60 vs. .51; the differences are largest for \emotion{joy} (.76 vs. .58), \emotion{surprise} (.66 vs. .50), and \emotion{disgust} (.50 vs. .36). For appraisal regression, both approaches perform similarly -- overall RMSE is 1.31 vs. 1.35 (text-only) and 1.34 vs. 1.36 (multimodal) -- though \emotion{pleasantness} (1.03 vs. 1.19) and \emotion{unpleasantness} (1.17 vs. 1.26) show larger differences. The limited appraisal difference may reflect that event descriptions capture what occurred rather than the subjective interpretation of the event, which is central to appraisals.

The difference in performance likely reflects the difficulty of reconstructing emotions from incomplete or ambiguous textual information (post text) versus more complete curated event descriptions. Overall, these results confirm that information lost during writing contributes to the difficulty of reconstructing authors' emotion processes.

\subsection{RQ3: How does image type and relevance impact emotion reconstruction?}
\label{sec:RQ3}
A key aspect of multimodal posts is the use of images, and how they interact with the text. We examine how image content and relationship to text vary by emotion and if performance for readers and models differs based on these factors.

\begin{figure}[t]
	\centering
	\includegraphics[width=.9\linewidth]{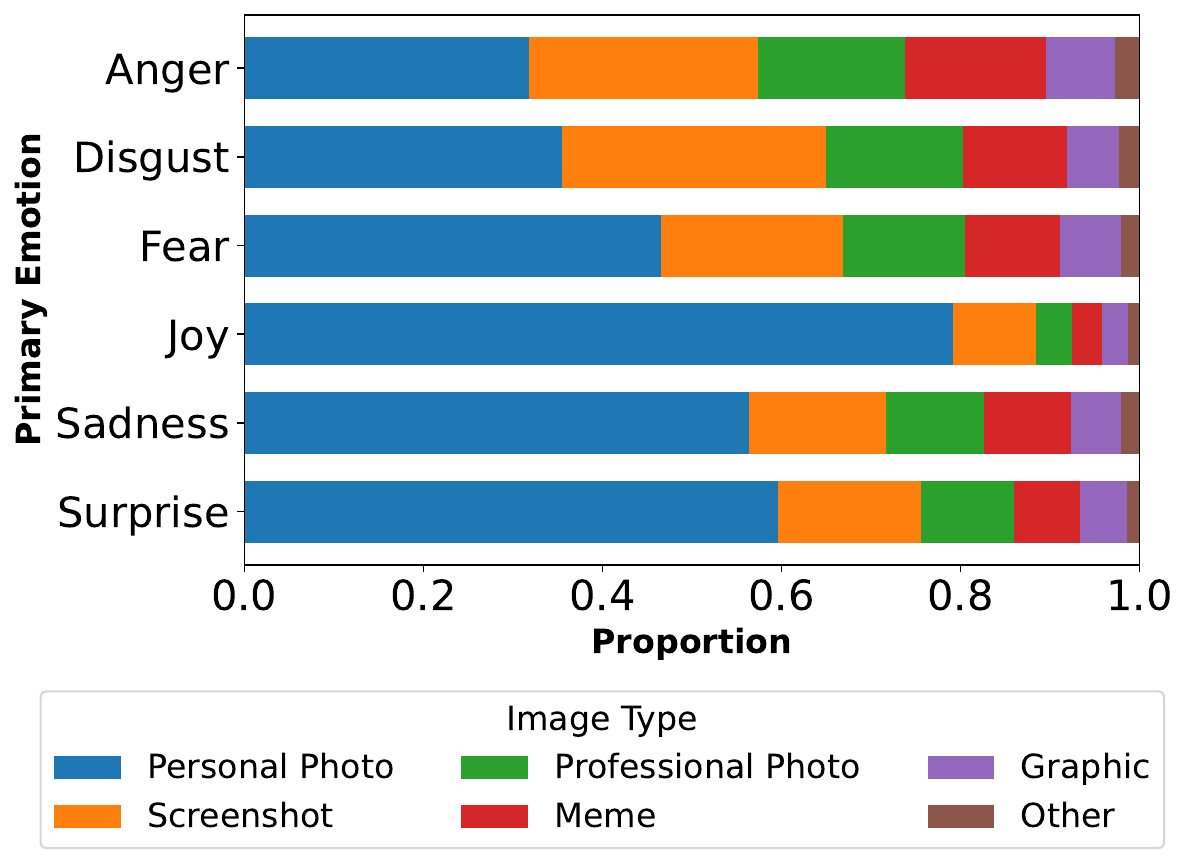}	
	\caption{Distribution of image type labels assigned by authors across primary post emotions.}
	\label{img_type_dist_fig}
\end{figure}

\paragraph{Experimental Setup.} 
We look at the distribution of image types (descriptions of image types in Appendix~\ref{App:Dataset_collection}), how authors and readers report the relationship between post text and image, and compare performance across these factors.
\paragraph{Results.}
Figure~\ref{img_type_dist_fig} shows that personal photos (PP) dominate across emotions, though \emotion{joy} posts use them more, while \emotion{anger}, \emotion{disgust}, and \emotion{fear} lean towards memes (M) and screenshots (SS). Table~\ref{tab:imglabel_f1_by_emotion} shows corresponding F1 scores: for \emotion{joy}, readers perform better on PP than memes, while for \emotion{fear} the reverse holds. Despite distributional differences in the training set, model and reader performance is similarly varied across image types, suggesting models do not exploit image type as a shortcut. Both readers and models perform better when text explicitly expresses the emotion and when text is required to understand the image, reinforcing the centrality of text for emotion reconstruction.\footnote{Full details in Appendix~\ref{App: Image vs text}}

{\setlength{\tabcolsep}{2pt}
\begin{table}[t]
\small
\centering
\begin{tabular}{lcccccccccc}
\toprule
 & \multicolumn{5}{c}{Reader $\uparrow$} & \multicolumn{5}{c}{CLIP $\uparrow$} \\
 \cmidrule(r){2-6} \cmidrule(l){7-11}
 & PP & Pro & SS & M & G  & PP & Pro & SS & M & G \\
 \cmidrule(r){2-6} \cmidrule(l){7-11}
Anger & .43 & .43 & .41 & .49 & .56$^{\dagger}$ & .43 & .42 & .49 & .42 & .74$^{\dagger}$ \\
Disgust & .40 & .41 & .37 & .47 & .30$^{\dagger}$ & .34 & .36 & .40 & .32 & .47$^{\dagger}$ \\
Fear & .41 & .50 & .42 & .47 & .59$^{\dagger}$ & .53 & .63 & .61 & .50 & .48$^{\dagger}$ \\
Joy & .63 & .56$^{\dagger}$ & .48 & .31$^{\dagger}$ & .43$^{\dagger}$ & .67 & .52$^{\dagger}$ & .37 & .25$^{\dagger}$ & .25$^{\dagger}$ \\
Sadness & .65 & .69 & .49 & .60 & .51$^{\dagger}$ & .65 & .66 & .52 & .43 & .52$^{\dagger}$ \\
Surprise & .32 & .45 & .37 & .33$^{\dagger}$ & .42$^{\dagger}$ & .53 & .59 & .42 & .55$^{\dagger}$ & .50$^{\dagger}$ \\
\cmidrule(r){2-6} \cmidrule(l){7-11}
Mac. Avg. & .47 & .51 & .42 & .45 & .47 & .52 & .53 & .47 & .41 & .49 \\
\bottomrule
\end{tabular}
\caption{Per-emotion F1 by image label type for readers and CLIP. PP: Personal Photo, Pro: Professional Photo, SS: Screenshot, M: Meme, G: Graphic. Cells marked with $^{\dagger}$ have support < 20 posts.}
\label{tab:imglabel_f1_by_emotion}
\end{table}
}

\section{Conclusion}
Our experiments show that readers and models can reconstruct authors' intended emotions and, to a lesser degree, appraisals from multimodal social media posts, though models outperform readers, especially on appraisals. Our multimodal analysis confirms that while images carry emotion-relevant information, text is the dominant signal for both readers and models; performance is highest when text explicitly conveys the emotion and contextualizes the image, and our baseline models do not effectively fuse the two modalities, which points to a clear direction for future work. We show the relationship between post and triggering event is crucial: when experienced and expressed emotions diverge, performance drops, and readers' understanding of the triggering event is a strong predictor of reconstruction accuracy. Yet even accurate event reconstruction does not guarantee correct emotion reconstruction, showing that reader misperception reflects not just reader failure but the inherent ambiguity of emotion expression on social media.

\multemo, as the first dataset to capture experience, expression, and perception in multimodal social media posts with both author and reader annotations, provides a foundation for future work. Future work should make use of \multemo's rich annotations, many of which we do not analyze here, such as comparing author and reader emotion stimluli annotations or investigating readers' explanations for their emotion annotations. \multemo can thus serve as a valuable resource to further investigate the complex relationship between experience, expression, and perception in multimodal social media posts, and to further explore the role of images in emotion expression. 

\section*{Acknowledgements}
We thank all ARR reviewers for their thorough reviews and valuable suggestions. We also thank
our study participants for contributing to our
work. This work has been supported by the
Deutsche Forschungsgesellschaft (DFG) in the
project “User’s Choice of Images and Text to Express Emotions in Twitter and Reddit” (ITEM,
Project KL 2869/11-1, No. 513384754).
\section*{Limitations}
Our study's limitations stem from two factors: (1)~the collection of \multemo, and (2)~our approach to modeling. 

\multemo is limited in several aspects. First, our participant pool is restricted to native English speakers in the UK and Ireland, which limits the generalizability of our findings to other languages and cultural contexts, as emotion expression and appraisal are known to vary cross-culturally. Second, recruitment difficulties for negative emotions, particularly \emotion{disgust}, mean that the author pool for different emotions is not fully comparable; authors of \emotion{disgust} posts were recruited over a longer period and under more flexible participation conditions than authors of \emotion{joy} posts, which may introduce systematic differences. Third, our reader annotations are collected from a general population rather than from the same social or cultural groups as the authors, which prior work suggests may suppress reader performance \cite{li-etal-2025-third}. Finally, while our appraisal annotations provide a rich characterization of authors' emotion experience, appraisals are inherently subjective and retrospective, and may themselves be subject to recall bias and social desirability.\footnote{See Appendix~\ref{App:Dataset_collection} for additional details.}

Our modeling experiments are limited in several aspects. First, we chose to limit the use of LLMs when doing modeling experiments, testing only a single MLLM and only using a single set of prompts. We chose this for three reasons: (1)~to avoid confounding the results with the known issues of LLMs, such as unknown training data and sensitivity to prompts, (2)~previous work using similar data has shown that LLMs do not outperform smaller models on emotion classification tasks \cite{bagdon-etal-2025-donate}, and (3)~we strongly believe in the value of smaller models fine-tuned on the task for understanding the data and setting a strong baseline for future work. 
Second, our baseline models use a simple concatenation fusion strategy which, as our results confirm, does not effectively exploit the image modality; the performance gaps we report therefore represent a lower bound on what is achievable with more sophisticated multimodal architectures.

Future work should address these limitations by extending \multemo to additional languages and cultural contexts, investigating whether reader performance improves when readers share social or cultural background with authors, and developing multimodal fusion methods better suited to the asymmetric and emotion-dependent relationship between text and image that we document here. \multemo, as the first dataset to capture experience, expression, and perception in multimodal social media posts provides a foundation for this future work.

\section{Ethical Considerations}
This study was approved by the ethics review board at Otto-Friedrich-Universität Bamberg. All participants were informed about the data collection procedure and the intended use of the data prior to participation. Nevertheless, we reflect on several potential challenges in this work.
Although participants consented to data use, individual posts may compromise anonymity in ways participants did not anticipate. Additionally, the collected data may contain information about third parties who did not actively participate in the study. In light of these two concerns, we decided to share \multemo for research purposes only, upon request. The publicly released dataset is anonymized by blurring faces in personal photographs and removing names and other identifying information from text.

VS Code Copilot was used to assist in writing the code for data analysis, limited to debugging, documentation, refactoring, and code completion. Claude Sonnet 4.6 was used to assist in writing the paper, limited to grammar and style suggestions, rephrasing for clarity, and organization of ideas. It was only used for paraphrasing or polishing the author’s original content, and never for suggesting new content.



\bibliography{custom}

\appendix

\section{Additional Dataset Collection Details}
\label{App:Dataset_collection}
Here we provide additional details about the dataset collection process, including participant recruitment, annotation details, and post-processing steps.
\subsection{Collection}
\paragraph{Process Overview.}
Participants for both stages are recruited through Prolific and are
required to be active users of social media platforms such as
Twitter(X), Instagram, or Facebook. Recruitment is limited to native
English speakers in the UK and Ireland to control for cultural
differences in emotion expression and appraisals. Participants are
compensated at a rate of £4.50 per survey, with the expected time to
complete the survey of 30 minutes. Participants may
complete one survey per emotion per stage. Author phase studies are
conducted on Google Forms and reader phase studies are conducted on a
custom Streamlit webapp. Collection occurred between July 2025 and
February 2026.

After participants annotate
their posts, they answer questions about
demographic information, personality traits, and social media use.

A full list of annotation questions are in Table~\ref{tab:postquestions_full} for author surveys and in Table~\ref{tab:postquestions_readers} for reader surveys. Options for multiple-choice questions for both authors and readers are provided in Table~\ref{tab:survey_options}.

\paragraph{Post Requirements.}
The posts are original posts written by the participants; they might include shared content from other authors (retweets, etc.) but they clearly include original content from the participant. Images might have text in them, but only when the post text has additional unique content. Participants are asked to provide the text and image from the post by copying and pasting the text into a text box and uploading the image file.

\paragraph{Collection Challenges.}
Participation varied by emotion, with \emotion{joy} the most popular and \emotion{disgust} the least. \emotion{Joy} studies were completed within days of being posted, while \emotion{disgust}, \emotion{anger}, and \emotion{fear} studies took months. To compensate for this we did studies of 100 participants (300 posts) per emotion. We did not start new studies until all emotions had finished, giving us posts from similar time spans for all emotions. Then we employed two strategies (for all emotions except joy): (1) we created separate studies which only required one post to make it easier for participants to find posts which fit the criteria, and (2) we allowed participants to participate in each study a second time, as long as they provided a different posts.

\begin{table}[t]
	\centering
	\small
	\begin{tabular}{lcccc}
		\toprule
		& \multicolumn{2}{c}{Original Image} & \multicolumn{2}{c}{Anon. Image} \\
		& I & T + I & I & T + I \\
		\cmidrule(r){2-3} \cmidrule(l){4-5}
		Anger  & .27 & .45  & .30 & .48 \\
		Disgust  & .23 & .36  & .20 & .37 \\
		Fear & .32 & .56  & .28 & .55 \\
		Joy  & .41 & .58  & .37 & .58 \\
		Sadness  & .31 & .60  & .30 & .57 \\
		Surprise  & .25 & .50  & .23 & .48 \\
		\cmidrule(r){2-3} \cmidrule(l){4-5}
		Overall & .30 & .51  & .28 & .50 \\
		\bottomrule
	\end{tabular}
	\caption{Comparison of models trained on original vs. anonymized images for emotion prediction.}
	\label{tab:anonymization_effects}
\end{table}

\subsection{Post-processing}
We manually review every post to ensure they contain both text and image content and remove posts which do not meet our criteria. The most common reason for removal is that the post is missing one modality or the image is a screenshot of the submitted text (screenshots of other posts alongside original text content are allowed). We decided post-hoc to remove posts which contain graphic content, specifically those which contain graphic violence or sexual content, to ensure the dataset is appropriate for all annotators and users. This criteria was not made clear to participants so they were still paid for their contributions. Additionally, many (1,744) posts required cropping of the image to reduce the image to the content which is present in the original post, as many submissions provided screenshotted images which included content outside of the original post, such as platform interface elements, comments, or other posts. We did this manually to ensure that only irrelevant content was removed.
\subsubsection{Anonymization}
To protect participant privacy, we anonymized the publicly released images: faces in personal photos are blurred, and names and other identifying information are removed from the text. Because this obscures cues present in the original posts, it may affect model performance. We therefore rerun all models on the anonymized data, so that reported results reflect the performance users can expect when working with the publicly released version of \multemo. Table~\ref{tab:anonymization_effects} reports the comparison; performance is close across both versions, suggesting that anonymization has little effect on model behavior. This result supports our finding that models rely heavily on the textual modality.
\subsection{Example Posts}
\begin{figure*}[!h]
  \centering
\begin{subfigure}{0.47\linewidth}
\includegraphics[width=\textwidth]{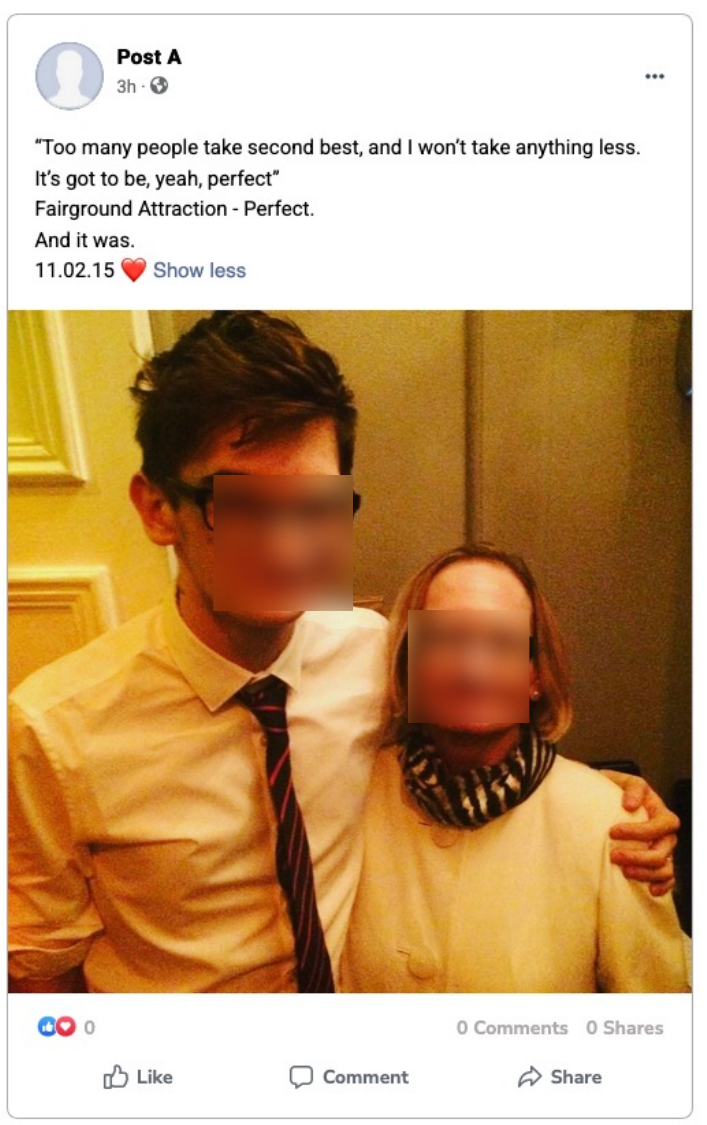}\caption{Post A: Sadness}
\label{fig:post_a}
\end{subfigure}
\hfill
\begin{subfigure}{0.47\linewidth}
\includegraphics[width=\textwidth]{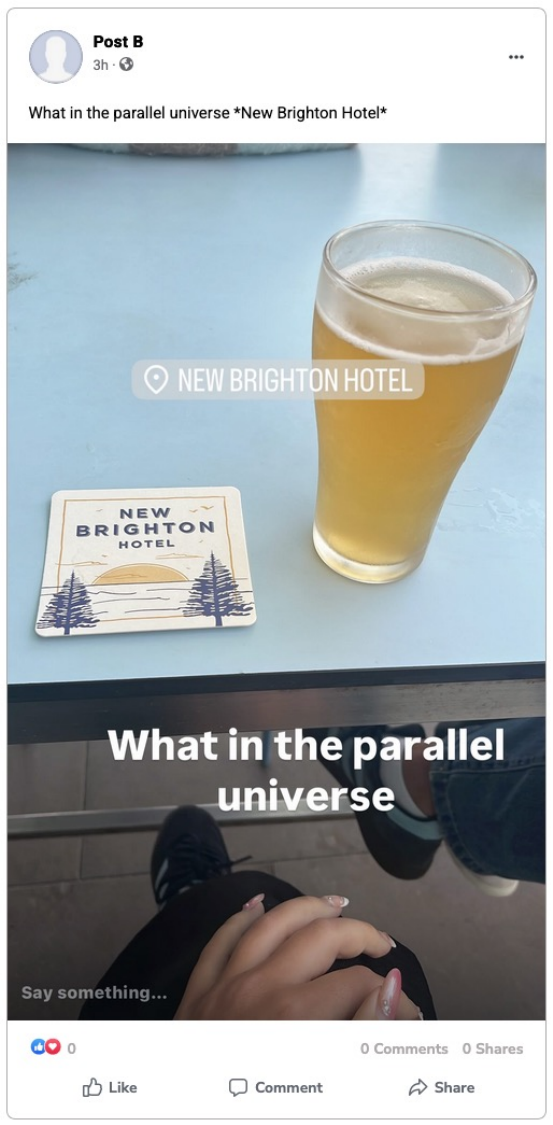}
\caption{Post B: Surprise}
\label{fig:post_b}
\end{subfigure}
\caption{Example posts from the dataset, including post text and image.}
\label{fig:example_posts}
\end{figure*}

In Figure~\ref{fig:example_posts} we show example posts from the dataset.

\begin{table*}[]
  \small
\begin{tabularx}{\textwidth}{llXr}
  \toprule
         & Emotion  & Event Description                                                                                                                             & Similarity Score \\
\cmidrule(r){1-1}\cmidrule(r){2-2}\cmidrule(r){3-3}\cmidrule(){4-4}
Post A & & &\\
Author   & Sadness  & It was to commemorate my mother who had passed away                                                                                           &                  \\
Reader 1 & Sadness  & I think maybe this post is about his mother/grandma  dying so im thinking the death of a loved one inspired him to create the post            & 0.55             \\
Reader 2 & Joy      & It looks like a son with his mother and the date it was taken                                                                                 & 0.14             \\
Reader 3 & Joy      & I really don`t have any idea , but  I`m guessing that the male did something that made the female proud                                       & 0.01 \\

\cmidrule(r){1-1}\cmidrule(r){2-2}\cmidrule(r){3-3}\cmidrule(){4-4}
Post B&&  & \\
Author   & Surprise & Because I was in Australia, and I went to a hotel which had the same name as the area I was raised in in the UK! It was a massive coincidence &                  \\
Reader 1 & Joy      & The author was enjoying an alcoholic beverage while staying in a nice hotel.                                                                  & 0.14             \\
Reader 2 & Surprise & The post is linking two things that are both named "New Brighton Hotel" - the actual hotel itself and a pint of beer.                         & 0.23             \\
Reader 3 & Surprise & They're surprised about something to do with the bar and the hotel I'd guess. What in the parallel universe indicates its surprising somehow. & 0.22             \\

\bottomrule    
\end{tabularx}
\caption{Author and reader event descriptions and their similarity scores for the example posts found in Fig.~\ref{fig:example_posts}.}
\label{tab:example_event_descriptions}
\end{table*}

\begin{table*}[p]
  \small
  \renewcommand{\arraystretch}{0.8}
  \begin{tabularx}{\linewidth}{p{3cm}Xl}
    \toprule
    Label & Question Text & Options \\
    \cmidrule(lr){1-1}\cmidrule(lr){2-2}\cmidrule(lr){3-3}
    \multicolumn{3}{l}{\textbf{Event Details}}  \vspace{.8ex} \\
        Event Description & Please describe the event which the post describes and your feelings about it by completing the following sentence: I felt sadness when/because/... Include event details or write multiple sentences if this helps us to understand the situation. & [Text]\\
        Event Duration & How long did the event last?  & [Time]\\
        Emotion Duration & How long did you experience emotion as a result
                       of the event?  & [Time]\\
        Event Recall &    How confident are you that you recall the event well? & 1\ldots5 \\ 
        Emotion (event) & Please select the primary emotion that you felt as a result of this event.  & [Emo.]\\
        Intensity & Please rate how intensely you felt each of these emotions as a result of this event. [Emo.] & [Inten.]\\
        
    \cmidrule(lr){1-1}\cmidrule(lr){2-2}\cmidrule(lr){3-3}
        \multicolumn{3}{p{14cm}}{\textbf{Appraisal:} Think back to when the event happened and recall its details. Take some time to remember it properly.
    How much do these statements apply? Some statements might not fit the event exactly, please answer to the best you can.}   \vspace{.8ex} 
    \\
    Suddenness & The event was sudden or abrupt. & 1\ldots5\\
    Familiarity & The event was familiar. & 1\ldots5\\
    Predictability & I could have predicted the occurrence of the event. & 1\ldots5\\
    Pleasantness & The event was pleasant for me. & 1\ldots5\\
    Unpleasantness & The event was unpleasant for me. & 1\ldots5\\
    Goalrelevance & I expected the event to have important consequences for me. & 1\ldots5\\
    Ownresponsibility & The event was caused by my own behavior. & 1\ldots5\\
    Otherresponsibility & The event was caused by someone else's behavior. & 1\ldots5\\
    Situationresponsibility & The event was caused by chance, special circumstances, or natural forces. & 1\ldots5\\
    Anticipconseq & I anticipated the consequences of the event. & 1\ldots5\\
    Goalsupport & I expected positive consequences for me. & 1\ldots5\\
    Urgency & The event required an immediate response. & 1\ldots5\\
    Owncontrol & I was able to influence what was occurring during the event. & 1\ldots5\\
    Otherscontrol & Someone other than me was influencing what was occuring. & 1\ldots5\\
    Chancecontrol & The event was the result of outside influences over which nobody had control. & 1\ldots5\\
    Acceptconseq & I anticipated that I would easily live with the unavoidable consequences of the event. & 1\ldots5\\
    Internalstandards & The event clashed with my standards and ideals. & 1\ldots5\\
    Externalstandards & The actions that produced the event violated laws or socially accepted norms. & 1\ldots5\\
    Attention & I had to pay attention to the situation. & 1\ldots5\\
    Notconsider & I tried to shut the situation out of my mind. & 1\ldots5\\
    Effort & The situation required me a great deal of energy to deal with it. & 1\ldots5\\
    
        \cmidrule(lr){1-1}\cmidrule(lr){2-2}\cmidrule(lr){3-3}
    \multicolumn{3}{l}{\textbf{Post Details}}   \vspace{.8ex}  \\
    Emotion & Considering both your image and what you wrote, please select the emotions that are present in your post. [multiple] & [Emo.]\\
    Intensity & Please rate how intensely you felt each of these emotions as a result of this event. &[Inten.]\\
    Emo. Stimulus (text) & Is the cause of the emotion(s) you felt described in the text of your post? If the cause is described in the text, please copy the text describing the cause and paste it here. & [Text]\\

    \cmidrule(lr){1-1}\cmidrule(lr){2-2}\cmidrule(lr){3-3}
    \multicolumn{3}{l}{\textbf{Image Details} }   \vspace{.8ex} \\
    Image Description & Please describe the content of the image by completing the following sentence: This is an image of ... & [Text]\\
    Image Type & Please select the description which best fits your image. If multiple choices apply to your image, please select the top-most description. & [Img-T]\\
    Image Reason & Why did you include an image alongside your text? [multiple]& [Img-R]\\
    Emo. Stimulus (image) & Is the cause of the emotion(s) you felt depicted in the image? If yes, please briefly describe it (i.e. "the dog on the right", "my grandma, the woman in the center of the group", or "the graduation ceremony I participated in".) & [Text]\\  

    \multicolumn{3}{l}{\textbf{Text--image relationship:}  How much do these statements apply? }   \vspace{.8ex} \\
    Text describes image & The text directly describes the image.  &
                                                                     1\ldots5
    \\
    Text $\rightarrow$ image & The text is required to understand
                               the image.  & 1\ldots5
    \\
    Image $\rightarrow$ text & The image is required to understand
                               the text.  & 1\ldots5
    \\
    Image conveys emotion & The image explicitly conveys the emotion you
                            posted about.  & 1\ldots5
    \\
    Text conveys emotion & The text explicitly conveys the emotion you
                           posted about.  & 1\ldots5
    \\
    \cmidrule(lr){1-1}\cmidrule(lr){2-2}\cmidrule(lr){3-3}
    \multicolumn{3}{l}{\textbf{Final Post Questions} }   \vspace{.8ex} \\
    Context & Does the post require additional context to understand? Select all that apply. & [Context]\\
    Reason for posting & Why did you make this post? (Select all that apply) & [Reason]\\
    Audience & Who did you intend to reach with this post? (Select all which apply) & [Audience]\\

    \bottomrule
  \end{tabularx}
  \caption{Author Phase: Wording and response options for survey questions used in the analysis.[Text] refers to free text responses. [Emo.] refers to Anger, Disgust, Fear, Joy, Sadness,
    Surprise. [Time] refers to one of Seconds, Minutes, Days, Weeks, Months. [Inten.], [Img-T], [Img-R], [Context], [Reason], and [Audience] refer to multiple choice options which can be found in Table~\ref{tab:survey_options}.}
  \label{tab:postquestions_full}
\end{table*}

\begin{table*}[!ht]
  \small
  \renewcommand{\arraystretch}{0.8}
  \begin{tabularx}{\linewidth}{p{3cm}Xl}
    \toprule
    Label & Question Text & Options \\
    \cmidrule(lr){1-1}\cmidrule(lr){2-2}\cmidrule(lr){3-3}
    \multicolumn{3}{l}{\textbf{Event Details}}  \vspace{.8ex} \\
        Event Description & Please describe the event which inspired the author to create this post. If the event is not obvious from the post, please make your best guess. You will be asked more questions about the event later in the survey. & [Text]\\
       
        Emotion (event) & What emotion(s) do you think the author felt when experiencing this event? (Please note: this may be different from the emotion they expressed in the post itself)  & [Emo.]\\
        Intensity & Please rate the intensity of each emotion you believe the author experienced during the event. (Please note: this may be different from the emotion they expressed in the post itself) [Emo.] & [Inten.]\\
        Explanation (event) & Please explain why you believe this to be the primary emotion.& [Text]\\
        Confidence (event) & How confident are you in your selection of the primary emotion?How confident are you in your selection of the primary emotion? & [Con.] \\

    \cmidrule(lr){1-1}\cmidrule(lr){2-2}\cmidrule(lr){3-3}
        \multicolumn{3}{p{14cm}}{\textbf{Appraisal:} Given the event you described from the post, how much do each of the following statements apply? The author of the post answered the same questions. Please try your best to guess the same answer as the post's author. Some statements might not apply to the event, but please rate them to the best of your ability.}   \vspace{.8ex} 
    \\
    Suddenness & The event was sudden or abrupt. & 1\ldots5\\
    Familiarity & The event was familiar to the author. & 1\ldots5\\
    Predictability & The author could have predicted the occurrence of the event. & 1\ldots5\\
    Pleasantness & The event was pleasant for the author. & 1\ldots5\\
    Unpleasantness & The event was unpleasant for the author. & 1\ldots5\\
    Goalrelevance & The author expected the event to have important consequences for themself. & 1\ldots5\\
    Ownresponsibility & The event was caused by the author's own behavior. & 1\ldots5\\
    Otherresponsibility & The event was caused by someone else's behavior. & 1\ldots5\\
    Situationresponsibility & The event was caused by chance, special circumstances, or natural forces. & 1\ldots5\\
    Anticipconseq & The author anticipated the consequences of the event. & 1\ldots5\\
    Goalsupport & The author expected positive consequences for themself. & 1\ldots5\\
    Urgency & The event required an immediate response. & 1\ldots5\\
    Owncontrol & The author was able to influence what was occurring during the event. & 1\ldots5\\
    Otherscontrol & Someone other than the author was influencing what was occuring. & 1\ldots5\\
    Chancecontrol & The event was the result of outside influences over which nobody had control. & 1\ldots5\\
    Acceptconseq & The author anticipated that they would easily live with the unavoidable consequences of the event. & 1\ldots5\\
    Internalstandards & The event clashed with the author's standards and ideals. & 1\ldots5\\
    Externalstandards & The actions that produced the event violated laws or socially accepted norms. & 1\ldots5\\
    Attention & The author had to pay attention to the situation. & 1\ldots5\\
    Notconsider & The author tried to shut the situation out of my mind. & 1\ldots5\\
    Effort & The situation required me a great deal of energy to deal with it. & 1\ldots5\\
    
        \cmidrule(lr){1-1}\cmidrule(lr){2-2}\cmidrule(lr){3-3}
    \multicolumn{3}{l}{\textbf{Post Details}}   \vspace{.8ex}  \\
    Emotion & What emotion(s) do you think the author of the post is expressing in the post itself? Please consider both the text and the image of the post.
     [multiple] & [Emo.]\\
    Intensity & Please rate the intensity of each emotion you believe the author expressed in the post
     &[Inten.]\\
     Explanation (post)& Please explain why you believe this to be the primary emotion expressed by the author in the post. & [Text]\\
     Confidence (post) & How confident are you in your selection of the primary emotion? & [Con.] \\
    Emo. Stimulus (text) &Does the text of the post indicate what caused the author’s emotion? What part of the text tells you what caused the author’s emotion? You can copy and paste or type it out below. & [Text]\\

    \cmidrule(lr){1-1}\cmidrule(lr){2-2}\cmidrule(lr){3-3}
    \multicolumn{3}{l}{\textbf{Image Details} }   \vspace{.8ex} \\
    Image Description & Please describe the content of the image by completing the following sentence: "This is an image of\dots" & [Text]\\
    Image Type & Please select the description which best fits your image. If multiple choices apply to your image, please select the top-most description. & [Img-T]\\
    Emo. Stimulus (image) & Does the image indicate what caused the author’s emotion? If yes, What in the image shows what caused the author's emotion? & [Text]\\  

    \multicolumn{3}{l}{\textbf{Text--image relationship:}  Please rate the following statements based on how much you agree with them.
 }   \vspace{.8ex} \\
    Text describes image & The text directly describes the image.  &
                                                                     1\ldots5
    \\
    Text $\rightarrow$ image & The text is required to understand
                               the image.  & 1\ldots5
    \\
    Image $\rightarrow$ text & The image is required to understand
                               the text.  & 1\ldots5
    \\
    Image conveys emotion & The image explicitly conveys the emotion you
                            posted about.  & 1\ldots5
    \\
    Text conveys emotion & The text explicitly conveys the emotion you
                           posted about.  & 1\ldots5
    \\
    \cmidrule(lr){1-1}\cmidrule(lr){2-2}\cmidrule(lr){3-3}
    \multicolumn{3}{l}{\textbf{Final Post Questions} }   \vspace{.8ex} \\
    Context & Does the post require additional context to understand? Select all that apply. & [Context]\\
    Reason for posting & Why do you think the author posted this on social media? (Select all which apply)Why do you think the author posted this on social media? (Select all which apply) & [Reason]\\
    Audience & Who do you think the author intended to reach with this post? (Select all which apply) & [Audience]\\

    \bottomrule
  \end{tabularx}
  \caption{Reader Phase: Wording and response options for survey questions asked of readers annotating posts. [Text] refers to free text responses. [Emo.] refers to Anger, Disgust, Fear, Joy, Sadness,
    Surprise. [Time] refers to one of Seconds, Minutes, Days, Weeks, Months. [Inten.], [Img-T], [Img-R], [Context], [Reason], and [Audience] refer to multiple choice options which can be found in Table~\ref{tab:survey_options}.}
  \label{tab:postquestions_readers}
\end{table*}

\begin{table*}[t]
	\small
\begin{tabularx}{0.95\linewidth}{p{3cm}X}
  \toprule
  Label & Multiple Choice Options  \\
  \cmidrule(lr){1-1}\cmidrule(lr){2-2}
 
  Emotion & \emotion{Anger}, \emotion{Disgust}, \emotion{Fear}, \emotion{Joy}, \emotion{Sadness}, \emotion{Surprise}  \\
  Intensities & Very slightly or not at all, A little, Moderately, Quite a bit, Extremely \\
  Image Type & Meme (A graphic or photo with text overlayed, often from a prescribed format),\\ 
  &Screenshot (Taken on computer, phone or tablet), \\ 
  &Graphic (Painting or drawing (digital or physical), photoshoped photo, etc), \\ 
  &Professional Photo (from news, sports, stock photo, marketing, etc), \\ 
  &Personal Photo (taken for personal reasons by you or someone else), \\ 
  &Other (No categories fit your image)  \\
  Image Reason & The image communicates the content more clearly and/or quickly than text. \\
  & The image is the focus of the post.\\
  & Posts with images receive more engagement.\\
  & To better attract attention. \\
  & I prefer making posts with images. \\
  & To trigger an emotion in the readers. \\
  Reason to Post & To advocate for something or someone (a figure, movement, idea, etc) or to convince people of something.\\
  & To promote events, products, organizations etc.\\
  & To communicate something about their personal life, for example using selfies, pictures of belongings (e.g., pets, clothes), etc.\\
  &  To express emotion, attachment, or admiration at an external entity or group.\\
  & To relay information regarding a subject or event using factual language.\\
  & To entertain using art, humor, memes, etc.\\
  & To directly attack an individual or group.\\
  & To be shocking or controversial.\\
  Audience & Friends, Family, Coworkers, Customers or clients, Followers or fans, Strangers \\
  Context & The post requires knowledge of specific event (such as the pandemic or an election).\\
  & The post requires knowledge of a specific group (such as a sports team or organization).\\
  & The post requires knowledge of a specific location (such as a city or country).\\
  & The post requires knowledge of a specific culture (such as a cultural practice or tradition).\\
  \bottomrule
\end{tabularx}

\caption{Multiple choice options for survey questions.}
\label{tab:survey_options}
\end{table*}

\clearpage

\section{Additional Dataset Analysis}
\label{App: Dataset Analysis}
\label{sec:dataset_analysis_appendix}
In this section we provide additional analysis of the \multemo dataset, including

\subsection{Emotion Content of Posts}
We analyze the emotion content of posts across emotions, including the use of explicit emotion words and the presence of multiple emotions. These factors might contribute to differences in how easily readers can reconstruct authors' intended emotion expression across emotions.
\subsubsection{Emotion Words in Post Text}
Post text can express emotion implicitly or explicitly through the use of emotion words, which can lead to differences in how easily readers can reconstruct authors' intended emotion expression. While explicit emotion words of the primary emotion might make it easier for readers to reconstruct the author's expressed emotion, the presence of emotion words of other emotions might lead to confusion. Differences in the use of emotion words across emotions might explain differences in how easily readers can reconstruct authors' intended emotion expression across emotions.

\begin{figure}[t]
	\centering
	\includegraphics[width=\linewidth]{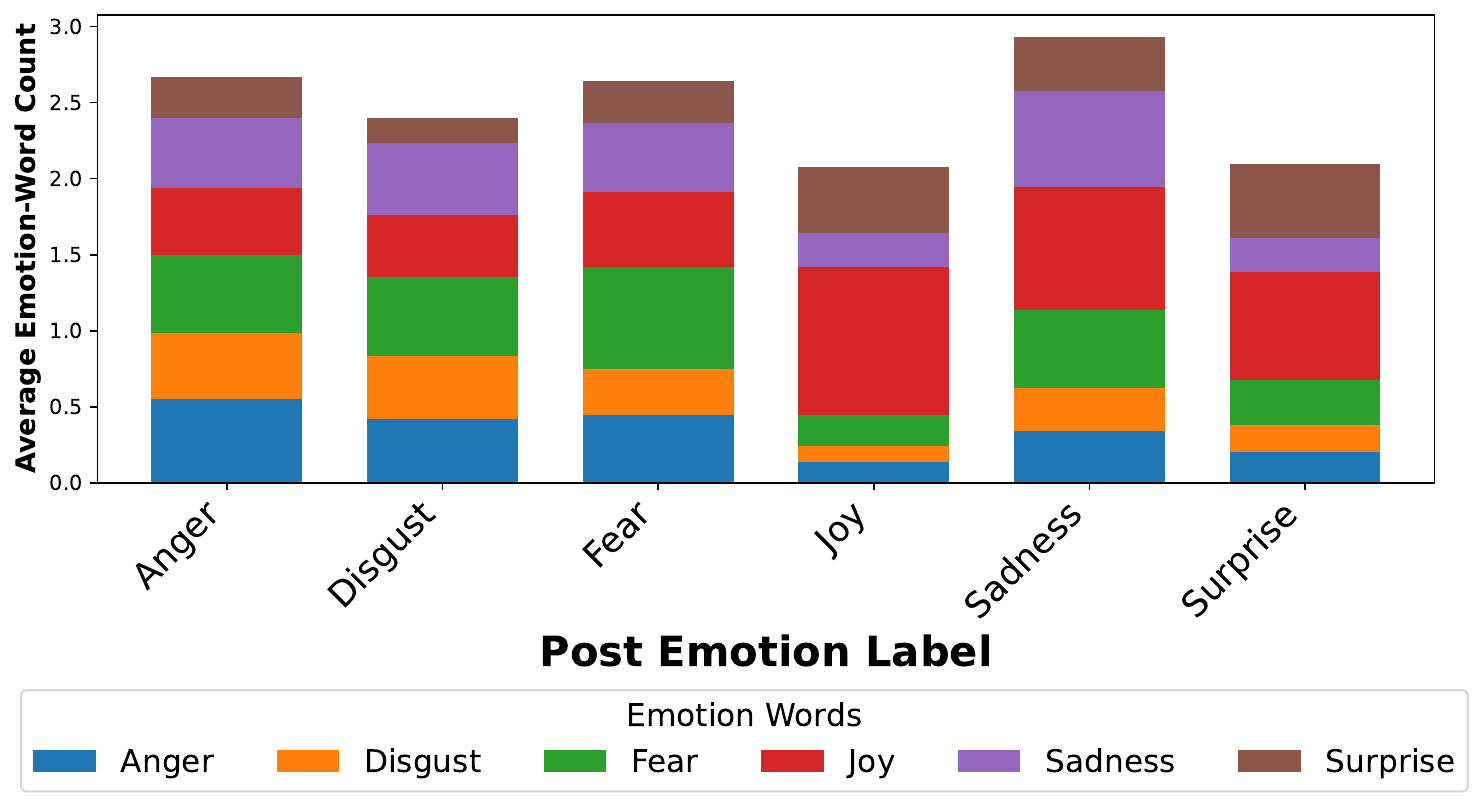}
	\caption{Average count of emotion words in post text for each emotion. Emotion words are identified using the NRC Emotion Lexicon.}
	\label{emotion_word_proportions}
\end{figure}

We count the number of emotion words in
the post text using the NRC Emotion Lexicon
\cite{EmotionDynamics,EmotionDynamics2}. We report the average
count of emotion words in the post text for each emotion, shown in Figure~\ref{emotion_word_proportions}, and find the use of explicit emotion words varies by primary post emotion. While
\emotion{joy} posts have a lower total emotion word count per post, the emotion words in \emotion{joy} posts are most likely to be \emotion{joy} words, while the emotion words in other emotion posts are more likely to be a mix of different emotion words. \emotion{Anger} and \emotion{disgust} posts have more even distributions of various emotion words, while \emotion{sadness} has the most emotion words per post on average, including a high number of \emotion{joy} words. All of this adds to the complexity of reconstructing authors' intended emotion expression, as readers might be confused by the presence of emotion words of other emotions, making it difficult to determine the primary emotion.

\subsubsection{Multiple Emotions}
\label{App: Multiple Emotions}


As we saw above, authors often use emotion words for multiple emotions, however this might not always translate to multiple emotions being expressed in a post. Here we investigate how often authors intend to express multiple emotions, which emotions are expressed together, and if readers perform differently based on these differences. 
\paragraph{Experimental Setup.}
We use authors' emotion intensity ratings to determine which emotions are expressed in posts, counting first the total number of emotions per post, and then counting how often emotions are expressed together. Then we analyze differences in reader and model performance based on the number of emotions expressed.

Additionaly, we use readers' emotion intensity ratings to determine if readers' secondary emotion choices are accurate. We rank readers' choices of primary emotion by their intensity ratings (i.e. the emotion with the highest intensity rating, excluding their primary choice, is considered the reader's secondary emotion choice, the second highest is their tertiary emotion choice, etc.)

\paragraph{Results.}
\begin{figure}[t]
	\centering
	\includegraphics[width=\linewidth]{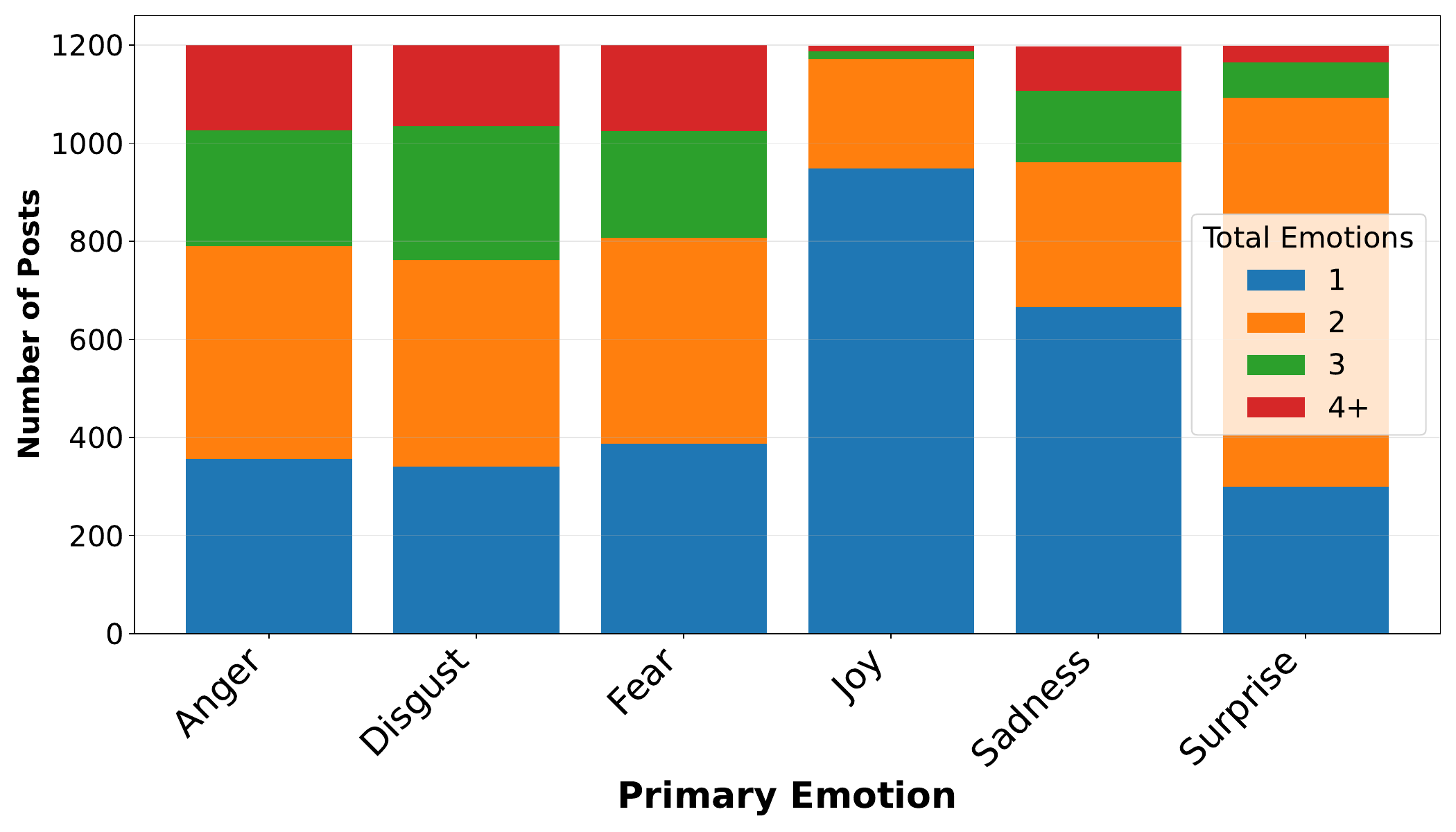}
	\caption{Counts of emotions expressed in posts by primary emotion, based on authors' emotion intensity ratings. }
	\label{emotion_cooccurring_stacked}
\end{figure}

\begin{figure}[t]
	\centering
	\includegraphics[width=\linewidth]{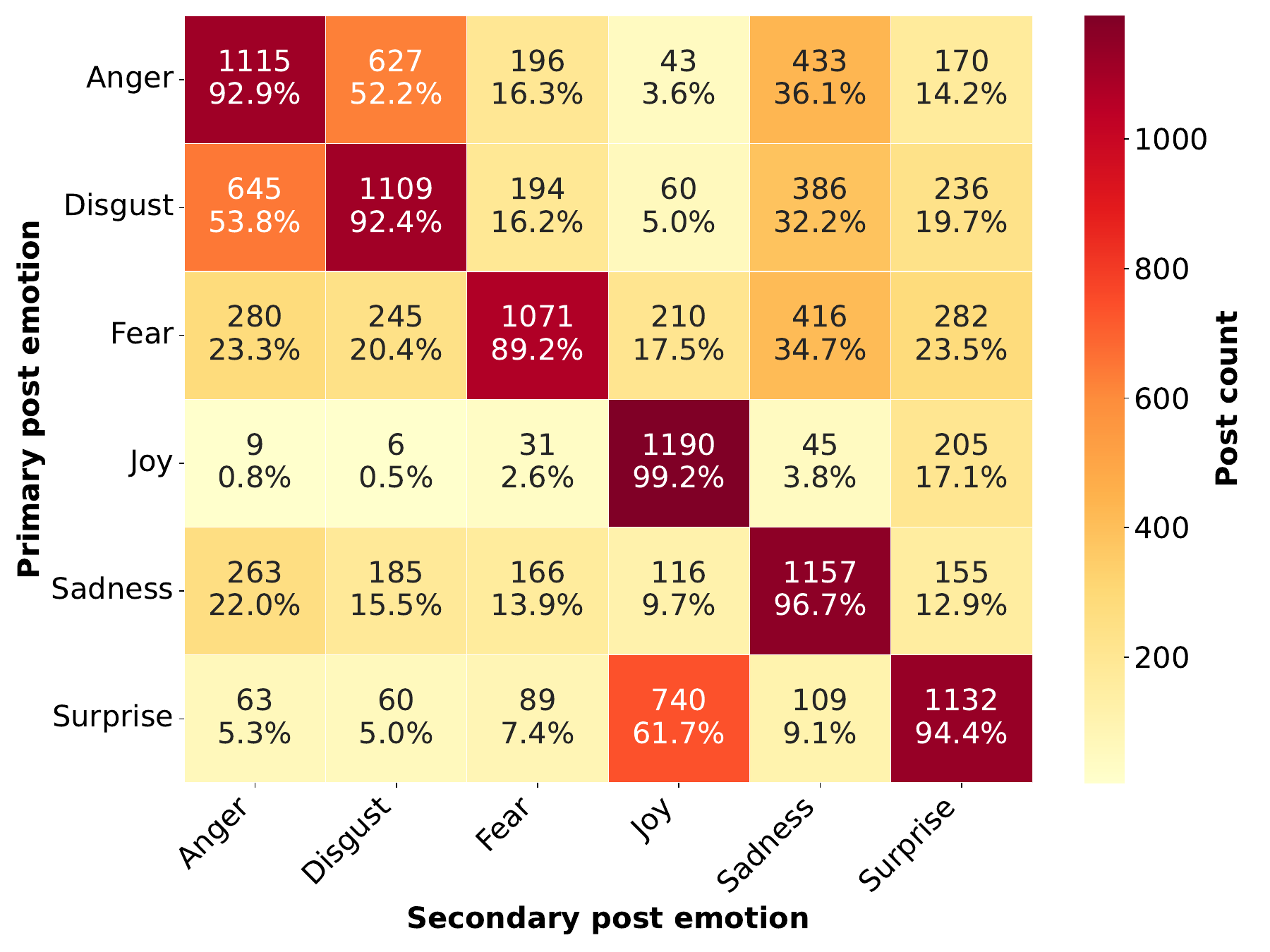}
	\caption{Co-occurring emotions heatmap showing the percentage of
		posts (1200 per emotion) with each primary emotion that also
		express secondary emotions.}
	\label{emotion_heatmap}
\end{figure}
{\setlength{\tabcolsep}{3pt}
\begin{table}[t]
  \small
	\centering
\begin{tabular}{lcccccccc}
\toprule
 & \multicolumn{4}{c}{Reader} & \multicolumn{4}{c}{CLIP} \\
  \cmidrule(lr){2-5} \cmidrule(lr){6-9}
 & 1 & 2 & 3 & 4+ & 1 & 2 & 3 & 4+ \\
  \cmidrule(lr){2-5} \cmidrule(lr){6-9}
Anger & .47 & .50 & .41 & .32 & .47 & .55 & .40 & .41 \\
Disgust & .38 & .44 & .31 & .32 & .32 & .43 & .26 & .34 \\
Fear & .47 & .46 & .39 & .33 & .55 & .61 & .53 & .61 \\
Joy & .74 & .29 & .16$^{\dagger}$ & .37$^{\dagger}$ & .72 & .33 & .22$^{\dagger}$ & .14$^{\dagger}$ \\
Sadness & .69 & .58 & .43 & .36 & .67 & .57 & .36 & .42 \\
Surprise & .36 & .38 & .31 & .00$^{\dagger}$ & .37 & .65 & .33 & .13$^{\dagger}$ \\
  \cmidrule(lr){2-5} \cmidrule(lr){6-9}
Macro Avg. & .52 & .44 & .33 & .28 & .52 & .53 & .35 & .34 \\
\bottomrule
\end{tabular}
	\caption{Reader F1 by author-labeled post-emotion count and primary emotion. Cells marked with $^{\dagger}$ have support < 50 annotations. }
	\label{tab:multi-emotion}
\end{table}
}

Figure~\ref{emotion_cooccurring_stacked} shows that the number of secondary emotions per post varies by primary emotion. Negative emotions are not only expressed with other emotions more often but also with a higher number of emotions than \emotion{joy}. \emotion{Surprise} is the most likely to have secondary emotions and it is most likely to have exactly one secondary emotion, which is most often \emotion{joy}. We show this in Figure~\ref{emotion_heatmap}, a heatmap of the percentage of posts for each primary emotion (y-axis) that also express secondary emotions (x-axis).  \emotion{Joy} and \emotion{surprise} have a lopsided relationship; 62\% of \emotion{surprise}  posts also express \emotion{joy}, but only 17\% of \emotion{joy} posts also express surprise, which may help to explain why \emotion{surprise} is the most difficult for readers to reconstruct. \emotion{Anger} and \emotion{disgust} have a different relationship; both are often expressed with each other, with each present in about 50\% of the other emotion's posts, while still appearing in other emotions' posts as well. 

To understand how this interacts with reader and model performance, we compare reader F1 scores by the number of emotions expressed in the post (including primary emotion), shown in Table~\ref{tab:multi-emotion}. Overall, reader performance decreases with increasing number of emotions, while model performance only decreases once three or more emotions are present. We see a pattern similar in performance as we do for emotion co-occurrence: \emotion{joy} and \emotion{surprise} are lopsided; performance drops from .74 and .72~F1 for posts with only one emotion to .29 and .33 for posts with two emotions, for readers and models respectively, however, \emotion{surprise} improves slightly for readers (.36 to .38) and greatly for models (.37 to .65). Whereas performance on \emotion{anger} and \emotion{disgust} are similar: both increase when there are two emotions, for both readers and models, but then drop for posts with three or more emotions. 

When readers' fail to reconstruct the author's expressed emotion, are they unable to discriminate the primary emotion from secondary emotions, or are they unable to recognize the primary emotion at all? We answer this by examining readers' emotion intensity ratings; For posts where readers fail to identify the primary emotion, we look at how many readers rated the gold primary emotion as having an intensity higher than 0, and how it ranked among the other emotions. We show the counts for this in Table~\ref{tab:reader_secondary_guess_counts}. 

We find for only 22\% of incorrect primary emotion annotations, readers recognize the primary emotion as being present in the post: 8\% are tied for highest intensity with their primary emotion selection (2nd Tie), and 13\% being the second highest intensity readers selected (2nd). For 78\% of incorrect primary emotion annotations, readers did not recognize the gold primary emotion as being present in the post at all, showing that difficulty with reconstructing the intended emotion is not just a matter of misidentifying the primary emotion, but fully failing to recognize the presence of the primary emotion in the post. 

Overall, we find there are patterns to how multiple emotions are expressed in posts, and the presence of multiple emotions in a post can lead to confusion for readers. Furthermore, reader confusion is often not just an inability to identify the primary emotion, but a full misunderstanding of the emotional content of the post.

\begin{table}[t]
	\centering
  \small
\begin{tabular}{lc|cccc}
\toprule
 & N & 2nd ( Tie) & 2nd & 3rd & Total \\
\midrule
Anger & 410 & 10\% & 12\% & 1\% & 23\% \\
Disgust & 492 & 11\% & 10\% & 1\% & 22\% \\
Fear & 498 & 5\% & 12\% & 1\% & 18\% \\
Joy & 60 & 8\% & 13\% & 0 & 21\% \\
Sadness & 248 & 5\% & 13\% & 1\% & 19\% \\
Surprise & 510 & 7\% & 15\% & 0 & 22\% \\
\midrule
Overall & 2218 & 8\% & 13\% & 1\% & 22\% \\
\bottomrule
\end{tabular}

	\caption{Reader secondary-guess counts by gold emotion. N is the count of incorrect primary emotion annotations. 2nd (Tie) is the percentage of incorrect annotations where readers rated the gold primary emotion as having the same intensity as their chosen primary emotion. 2nd and 3rd are the percentages where readers rated the gold primary emotion as their second and third highest rated emotions, respectively.  
  }
	\label{tab:reader_secondary_guess_counts}
\end{table}

\subsection{Event vs. Post}
\label{App: Event vs Post}
In this section we extend the analysis of differences between experienced and expressed emotions described in Section~\ref{sec:rq2}.

\subsubsection{Emotions Experienced vs. Expressed}
\label{App: Experienced vs Expressed emotion}

The experience of an emotion during an event may differ from the expression of that emotion in a post about the event. We analyze how frequently this occurs and if readers or models perform worse when it does.
\paragraph{Experimental Setup.}
Using posts in \multemo's test set, we look at differences in the number of emotions experienced vs. expressed, which we categorize into more emotions in the post than the event (More in Post) or less emotions in the post than the event (Less in Post). For each, we compare the difference in reader ($\Delta$R) and model ($\Delta$M) performance on the corresponding subset of posts 
to their performance on the whole test set in predicting the primary emotion.
\paragraph{Results.}
In Table~\ref{tab:post_event_more_less_comparison}, and as we describe in more detail below, we find that authors  frequently express fewer emotions in their posts than they experienced during the event.
However, both reader and model performance does not suffer from this. While it is uncommon for authors to express more emotions than they experienced, when they do, readers perform worse and model performance varies by emotion. 

Authors are more likely to omit experienced negative emotions from their posts (column Less), while it is unlikely for authors to include additional emotions (column More). Performance on posts with less secondary emotions is similar to the overall performance, which is surprising because one might expect higher performance when less noise from secondary emotions is present. However, we do see differences in performance when additional emotions are present, though to varying degrees per primary emotion. Performance on \emotion{joy} and \emotion{surprise} posts is lower, for both readers and models, while performance is higher for \emotion{sadness} and \emotion{anger} posts. However, we caution to draw conclusions from this due to the small sample sizes. 

{\setlength{\tabcolsep}{3pt}
\begin{table}
  \small
	\centering
\begin{tabular}{lrrrrrr}
\toprule
 &  \multicolumn{3}{c}{More in Post} & \multicolumn{3}{c}{Less in Post}  \\
 \cmidrule(r){2-4} \cmidrule(l){5-7} 
 &  \%P & $\Delta$R & $\Delta$M & \%P & $\Delta$R & $\Delta$M  \\
 \cmidrule(r){2-4} \cmidrule(l){5-7} 
Anger  & 2 & $-$.01 & .27 & 78 & .03 & .01 \\
Disgust & 3 & $-$.31 & $-$.02 & 69 & .01 & $-$.01 \\
Fear & 3 & $-$.08 & $-$.04 & 63 & .00 & $-$.02 \\
Joy & 1 & $-$.26 & $-$.20 & 43 & $-$.03 & $-$.03 \\
Sadness & 3 & .09 & .18 & 64 & .00 & .00 \\
Surprise & 2 & $-$.07 & $-$.17 & 40 & $-$.04 & $-$.12 \\
 \cmidrule(r){2-4} \cmidrule(l){5-7} 
Macro Avg. & 2 & $-$.11 & .00 & 60 & $-$.00 & $-$.03 \\
\bottomrule
\end{tabular}
	\caption{Comparison of post vs.~event emotions by primary emotion. More (Less) in Post are posts with more (less) emotions in the post compared to the event. \%P shows the percentage of posts in the full dataset. $\Delta$R / $\Delta$M are the difference between reader / model performance (F1), respectively, on the full test set and the subset.}
	\label{tab:post_event_more_less_comparison}
\end{table}
}

\subsubsection{Post Text vs. Event Description}
\label{App:Post Text vs Event}
In this section we extend the analysis presented in Section~\ref{sec:rq2_event} to analyze the relationship between post text and event descriptions.

\multemo's event description annotations allow us to compare the post text to the author's description of the event as well as to the readers' event description that they infer from the post text. 
This gives us insight into how well readers infer the event from the post.
 To measure their similarity, we score each combination of post text (P) and event description from both authors~($E_A$) and readers ~($E_R$) using a cross-encoder model~\cite{reimers-2019-sentence-bert} trained on the STS Benchmark~\cite{enevoldsen2025mmtebmassivemultilingualtext,muennighoff2022mteb}.

{\setlength{\tabcolsep}{3pt}
\begin{table}
\centering
\small
\begin{tabular}{lcccccccc}
\toprule
 & \multicolumn{2}{c}{P/$E_A$} & \multicolumn{2}{c}{P/$E_R$} & \multicolumn{2}{c}{$E_A$/$E_R$} & \multicolumn{2}{c}{$E_R$/$E_R$} \\
 \cmidrule(lr){2-3} \cmidrule(lr){4-5} \cmidrule(lr){6-7} \cmidrule(lr){8-9}
 & M & STD & M & STD & M & STD & M & STD \\
 \cmidrule(lr){2-3} \cmidrule(lr){4-5} \cmidrule(lr){6-7} \cmidrule(lr){8-9}
Anger & .27 & .23 & .29 & .20 & .26 & .18 & .37 & .20 \\
Disgust & .27 & .24 & .27 & .19 & .28 & .18 & .39 & .18 \\
Fear & .33 & .24 & .33 & .20 & .28 & .18 & .41 & .19 \\
Joy & .30 & .22 & .36 & .20 & .34 & .17 & .44 & .19 \\
Sadness & .28 & .23 & .31 & .18 & .34 & .19 & .44 & .18 \\
Surprise & .34 & .24 & .35 & .20 & .34 & .18 & .44 & .19 \\
 \cmidrule(lr){2-3} \cmidrule(lr){4-5} \cmidrule(lr){6-7} \cmidrule(lr){8-9}
Macro Avg. & .30 & .23 & .32 & .20 & .30 & .18 & .41 & .19 \\
\bottomrule
\end{tabular}
\caption{Mean (M) cosine similarity and standard deviation (STD) between post text (P) and author's event description (E$_A$), post text and readers' event descriptions~($E_R$),  author's event description and readers' event descriptions, and between readers' event descriptions.}
\label{table:post_event_similarity}
\end{table}
}

Readers' event descriptions are more similar to each other ($E_R$/$E_R$) than they are to the authors' event descriptions ($E_A$/$E_R$), as shown in Table~\ref{table:post_event_similarity}. This indicates that readers frequently come to a similar understanding of the event. It also aligns with our findings in Table~\ref{table:emotion_classification_simple}; readers agree with each other (.57~F1) more than they agree with authors (.47~F1) on emotion classification (column~RR vs R). Combined, these results suggest that readers have a consistent understanding of the post, but that this understanding is often different from the author's intended expression.

Comparing similarity scores P/$E_A$~(.30) and P/$E_R$~(.32), we see that authors omit event details. Readers' descriptions of the event are influenced by the post text, while authors' descriptions of the event can include information that is not present in the post text. This is supported by the higher standard deviation in post text and authors' event description similarity, showing there is more variability between post text content and authors' event descriptions. 


Overall, we find that readers' understanding of the event is consistent among themselves, and that their descriptions of the event are more similar to the post text than authors' descriptions of the event.

\subsubsection{Post Text and Image Similarity}
\label{Appendix:image_text_sim}
 To analyze how post images play into reader understanding of events we measure post text and image similarity using visual question answering \cite{hessel-etal-2021-clipscore}. We use the BLIP model \cite{blip} to answer the question ``Does this image match the following social media post text: "\{\}"? Answer yes or no.''. We use the probability of the answer being ``yes'' as the similarity score between the text and the image. We do the same for the question ``Does this image match the following event description: "\{\}"? Answer yes or no.'' to get similarity scores between the image and event descriptions. We then compare how these scores differ between post text and event descriptions, and how they relate to reader and model performance.

\begin{figure}
    \centering
    \includegraphics[width=\columnwidth]{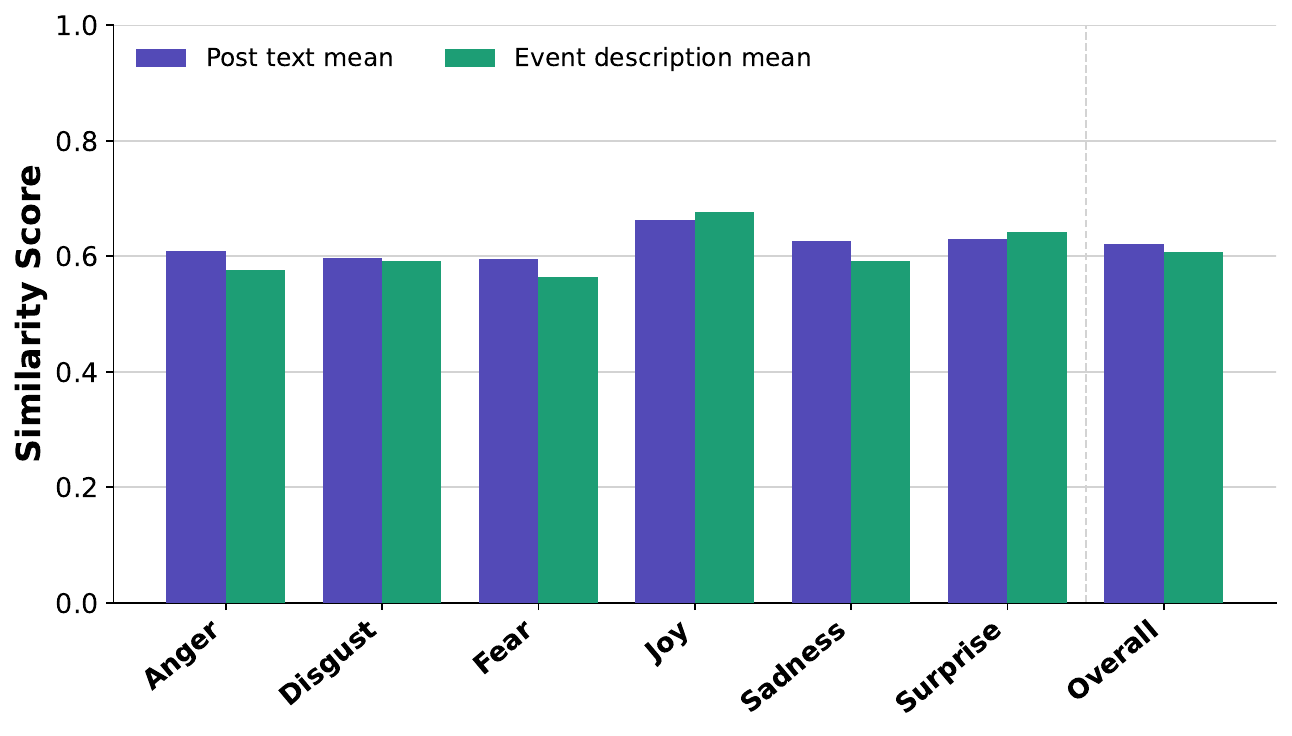}
    \caption{Similarity scores between post text and image and author event descriptions and images using VQA and BLIP.}
    \label{fig:post_image_similarity_comparison_bar_chart}
\end{figure}

In Figure~\ref{fig:post_image_similarity_comparison_bar_chart}, we find image similarity to post text and event descriptions varies little by emotion. \emotion{Joy} posts have slightly higher similarity between image and post text and event descriptions, while \emotion{disgust} \emotion{anger} and \emotion{fear} have slightly lower similarity, however these differences are small. 

\begin{figure}
    \centering
    \includegraphics[width=\columnwidth]{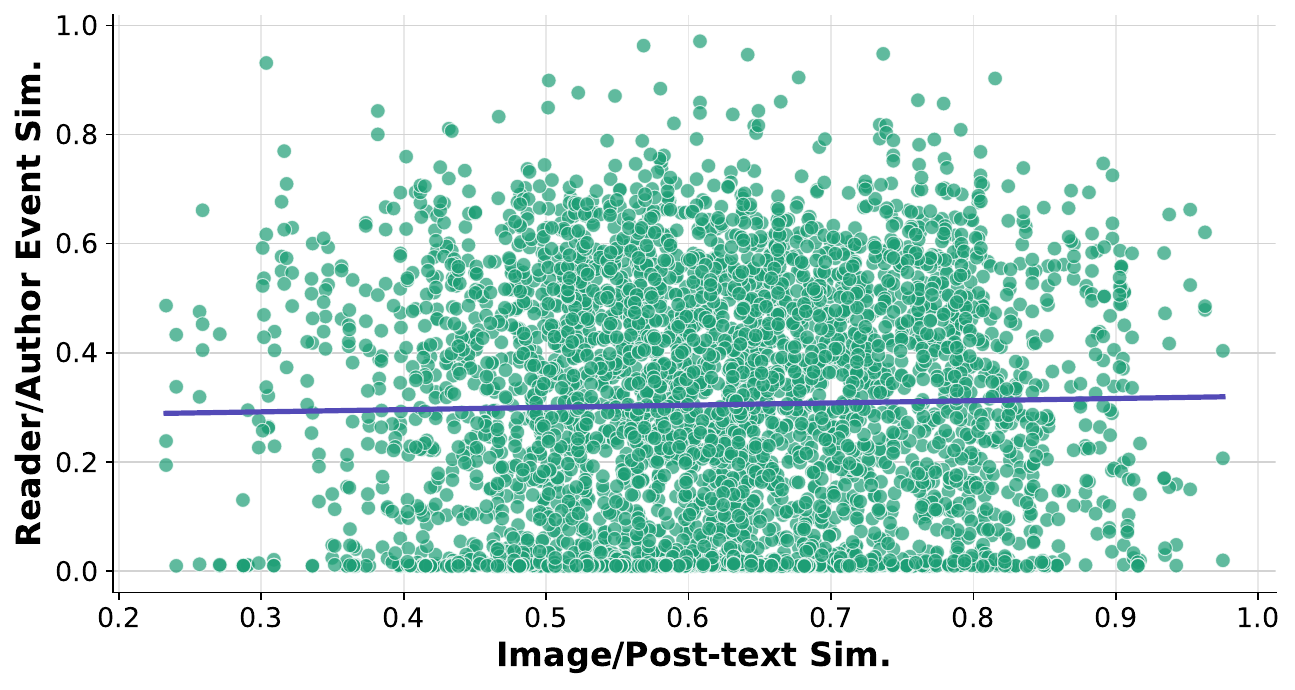}
    \caption{Scatter plot of post text and image similarity against reader and author event description similarity.}
    \label{fig:image_similarity_scatter_plot}
\end{figure}

In Figure~\ref{fig:image_similarity_scatter_plot} we compare post text and image similarity to readers' understanding of the event (Reader/Author event description similarity). We find that image similarity to post text does not correlate with readers' understanding of the event.This suggests that the images often do not contain information about the event that is not already present in the post text, and that the image is often not necessary for readers to understand the event. 

\subsection{Image--Text Relationship}
We extend the analysis of the relationship between post text and images presented in Section~\ref{sec:RQ3}.
\label{App: Image vs text}


\begin{figure*}
	\centering
	\includegraphics[width=\linewidth]{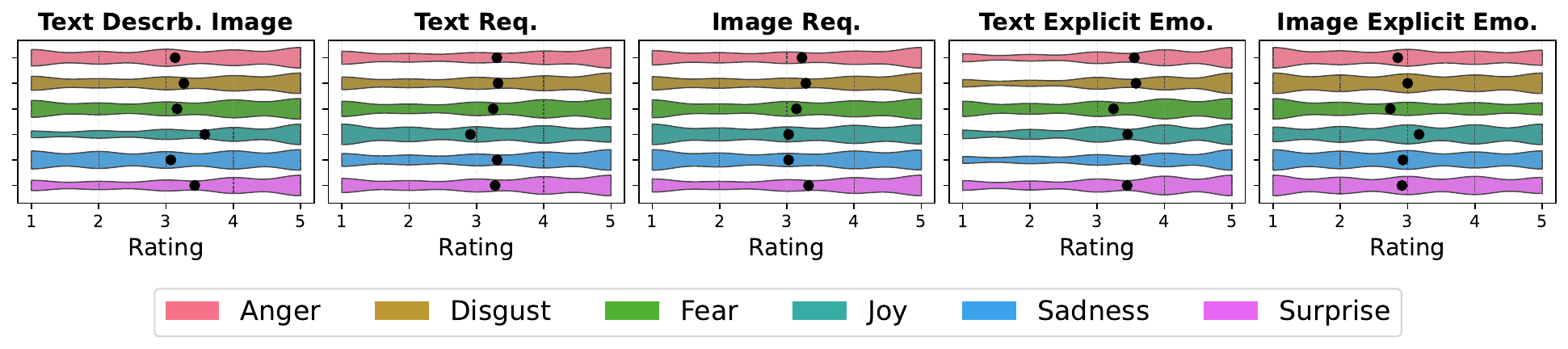}
	\caption{Distribution of image and text relationship ratings for each emotion. Emotions are labeled by color.}
	\label{img_text_relation}
\end{figure*}

We look at the distribution of image types across emotions, as labeled by authors, how authors and readers report the relationship between post text and image, and compare performance across these factors.
We find both the type of image and the relationship between image and text varies by primary post emotion and both readers and models perform differently based on these factors.

We can see the variance in image type labels assigned by authors in the main paper in Figure~\ref{img_type_dist_fig}. Personal photos are the most common image type for all emotions, but \emotion{joy} posts are more likely to use them than other emotions, while \emotion{anger}, \emotion{disgust}, and \emotion{fear} are more likely to use memes or screenshots than other emotions.

We report F1 scores per emotion for each image type in Table~\ref{tab:imglabel_f1_by_emotion} (main paper). Performance varies by image type, but the difference is greater within each emotion. For example, for \emotion{joy} posts, reader performance is much higher for personal photos than for memes, while for \emotion{fear} posts, reader performance is higher for memes than for personal photos. We considered that models might have learned to associate certain image types with certain emotions via the distribution of image types in the training set, but given how similar model performance is to reader performance across image types, this does not seem to be the case. 

Images' relationship with post text varies by emotion less than image type but we still observe differences in reader and model performance. In Figure~\ref{img_text_relation} we see  \emotion{joy} posts are more likely to directly describe the image, and their images more often explicitly depict the emotion, while also being less likely to require the text to understand the image. For all other emotions the text is more often required to understand the image than the opposite, though these differences are not large.

Differences in performance based on image--text relationship can be seen in Table~\ref{tab:text_img_rel}. 
We report reader performance by author annotation (A), reader annotation (R), and model performance by author annotation (CLIP). 
For both readers and models we find that performance is higher on posts whose texts explicitly express the emotion and where text is required to understand the image.
This shows that both, models and readers, rely heavily on the text to reconstruct the intended emotion. Furthermore, when readers perceive the text as describing the image they perform better, while there is little difference  in reader performance where authors report their text describes the image.

Our results here show that while differences in image type and relationship with the text show variance in reader and model performance, the results reinforce the importance of the text for reconstructing the intended emotion, as performance is higher when the text explicitly describes the emotion and when the text is required to understand the image while the opposite does not show the same pattern.


\begin{table}[t]
\centering
\small
\begin{tabular}{lrrrr|rr}
\toprule
& \multicolumn{4}{c|}{Readers} & \multicolumn{2}{c}{CLIP} \\
\cmidrule(lr){2-5} \cmidrule(lr){6-7}
 & \multicolumn{2}{c}{A. Anno.} & \multicolumn{2}{c|}{R. Anno.} &  \\

 & 1--2 & 3--5 & 1--2 & 3--5 & 1--2 & 3--5 \\
\midrule
T describes I & .47 & .46 & .43 & .49 & .49 & .52 \\
T required I & .44 & .48 & .43 & .48 & .48 & .53 \\
I required T & .48 & .45 & .49 & .45 & .52 & .51 \\
T explicit E & .41 & .48 & .32 & .49 & .44 & .54 \\
I explicit E & .45 & .47 & .46 & .47 & .52 & .51 \\
\bottomrule
\end{tabular}
\caption{\label{tab:text_img_rel} Performance (F1) of readers and models reported by relationship between text and image as annotated by authors (A) and readers (R). Text describes image (T des. I), text/image required to understand image/text (T/I req. I/T), text/image explicitly conveys emotions (T/I exp.).}
\end{table}

\begin{figure}[t]
	\centering
	\includegraphics[width=\linewidth]{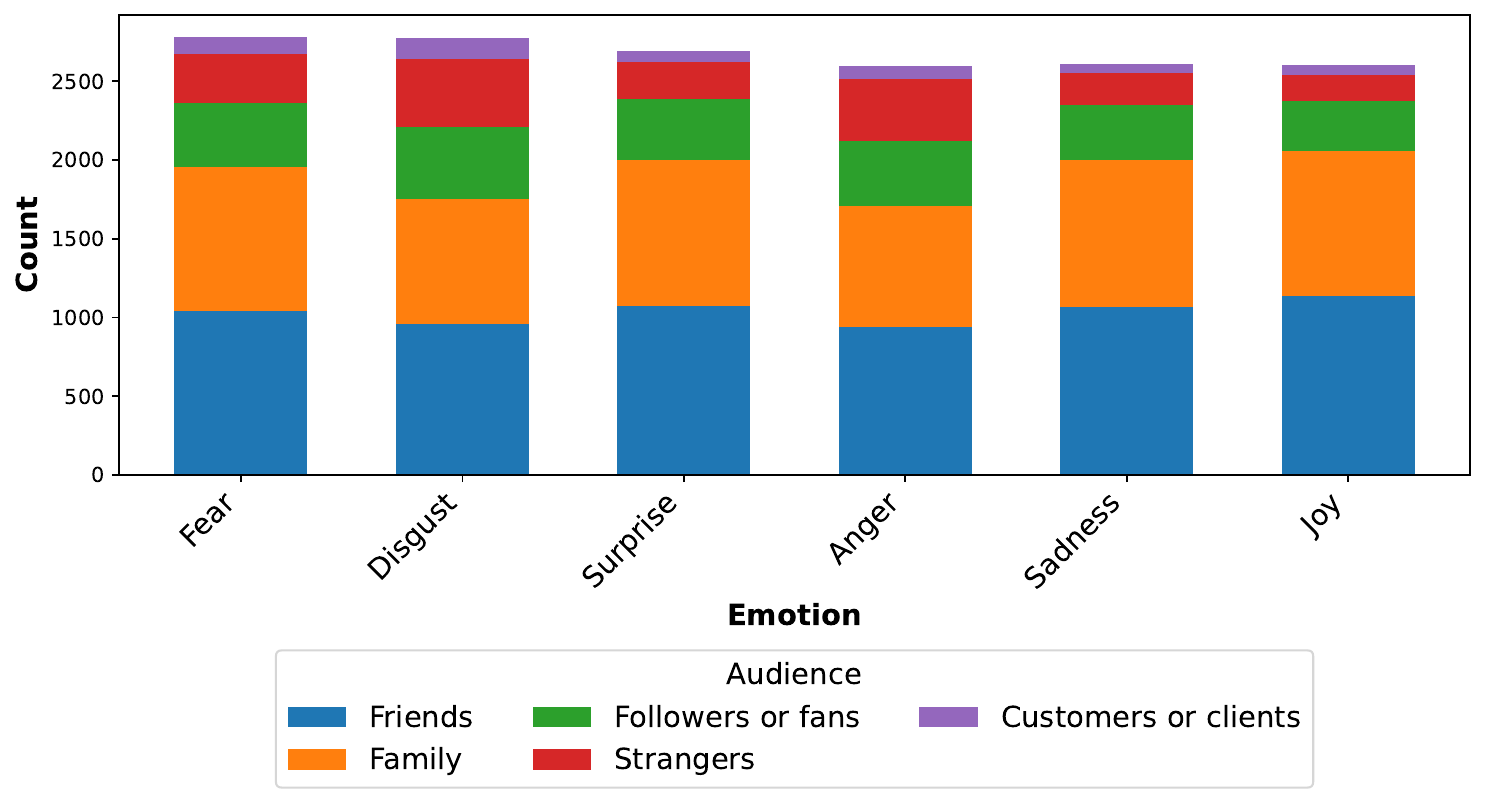}
	\caption{Counts of each target audience label for each emotion. Individual posts can have multiple target audience labels.}
	\label{post_audience_counts_fig}
\end{figure}

\begin{figure}[t]
	\centering
	\includegraphics[width=\linewidth]{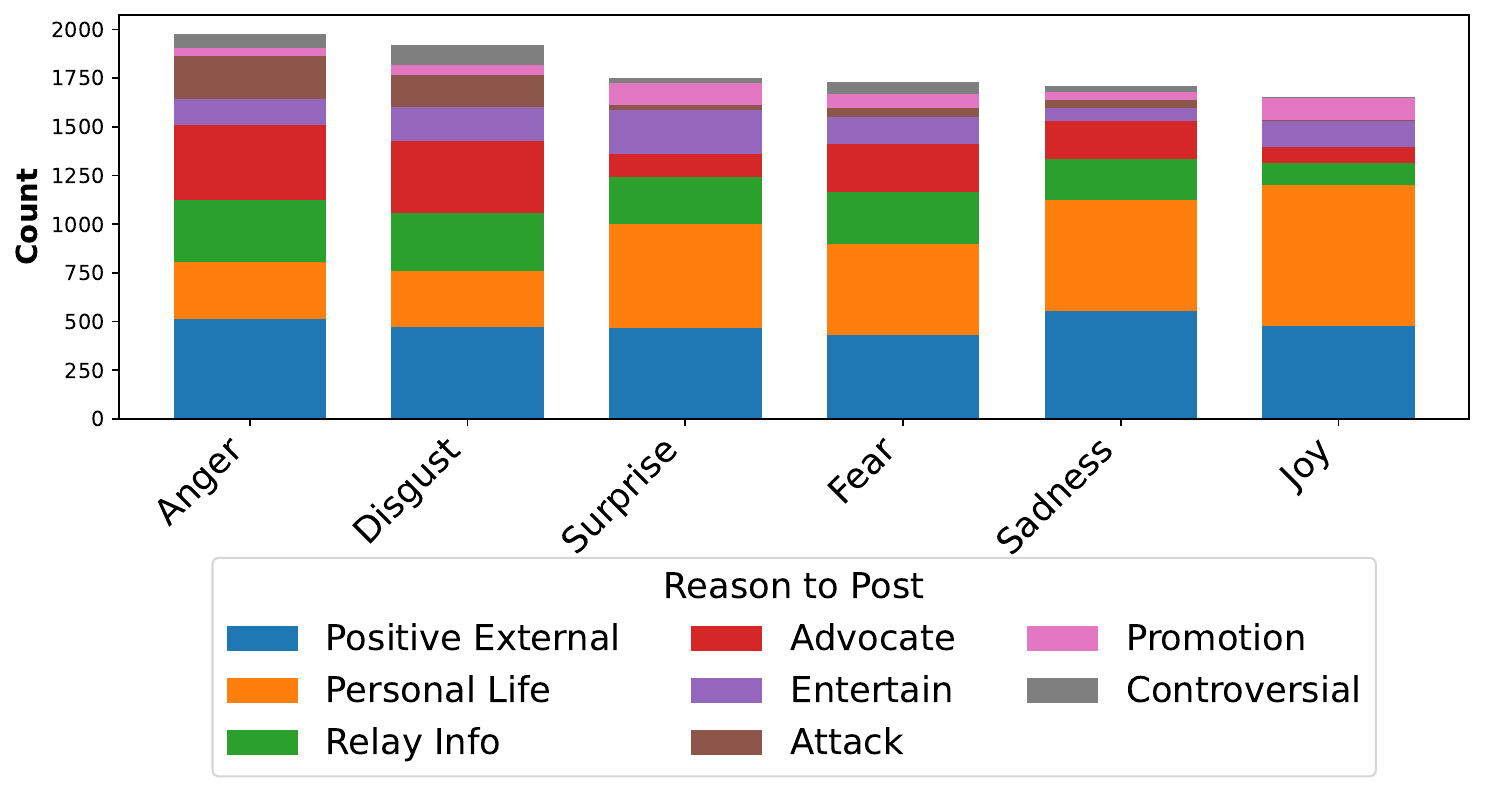}
	\caption{Authors' reported reasons for posting on social media, by emotion. Individual posts can have multiple reason labels.}
	\label{why_post_counts}
\end{figure}

\begin{figure}[t]
	\centering
	\includegraphics[width=\linewidth]{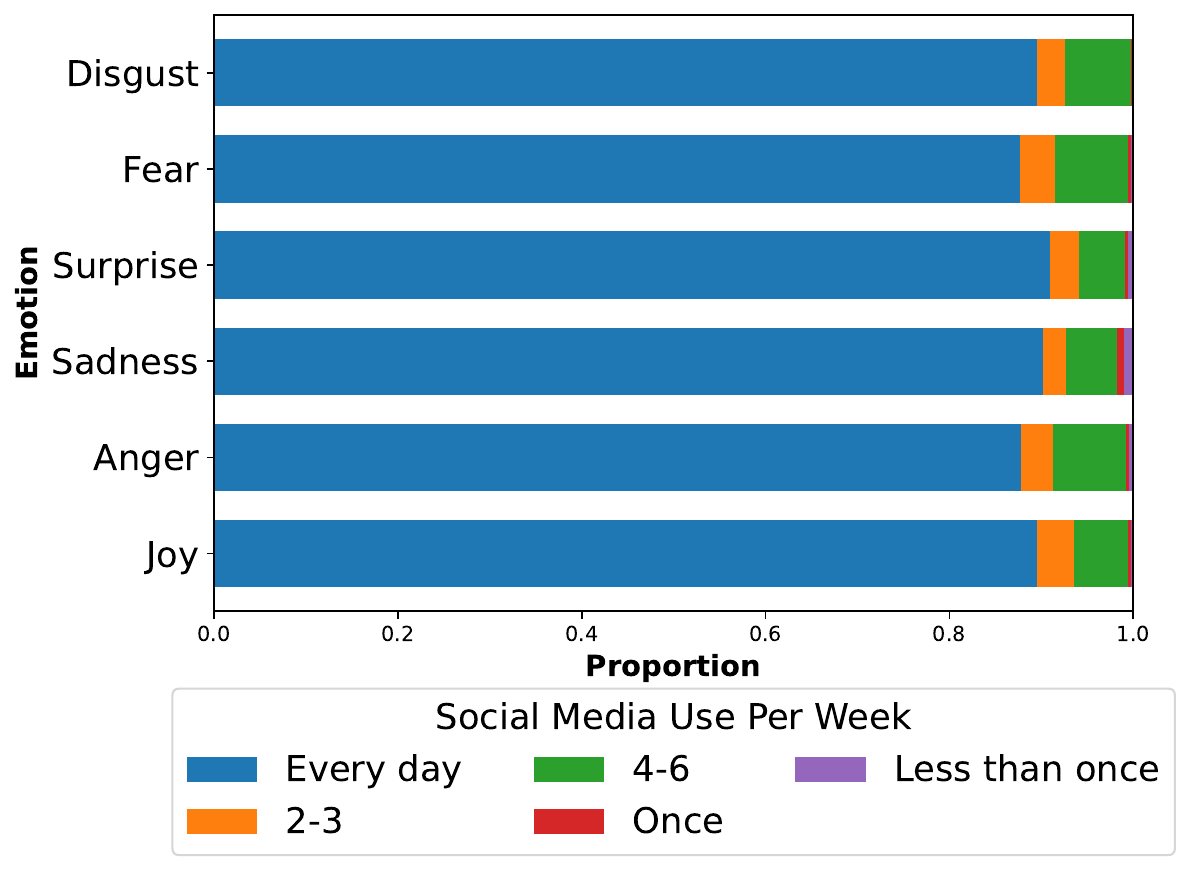}
	\caption{How often authors use (not post) social media, by emotion.}
	\label{social_media_use_proportions}
\end{figure}

\begin{figure}[t]
	\centering
	\includegraphics[width=\linewidth]{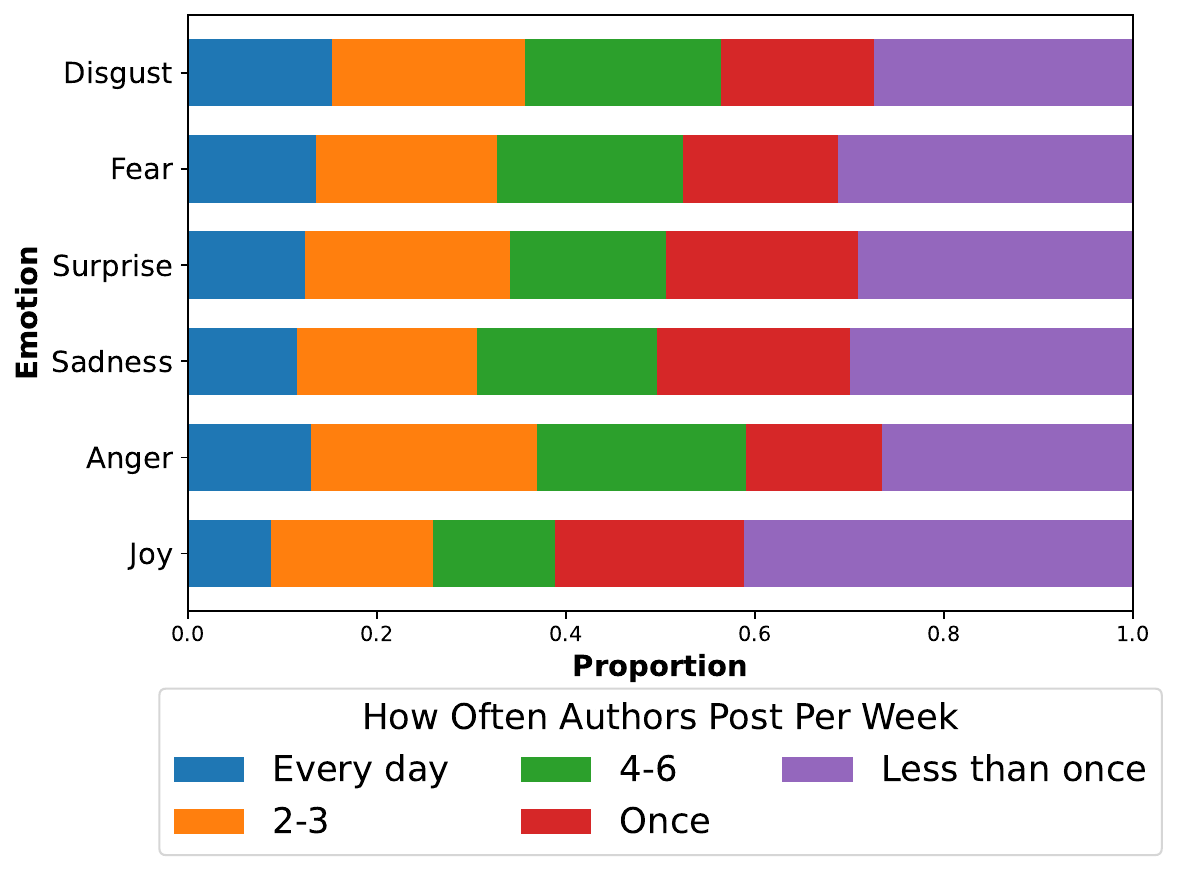}
	\caption{How often authors reported posting on social media, by emotion.}
	\label{social_media_post_freq_proportions}
\end{figure}

\begin{figure}[t]
	\centering
	\includegraphics[width=\linewidth]{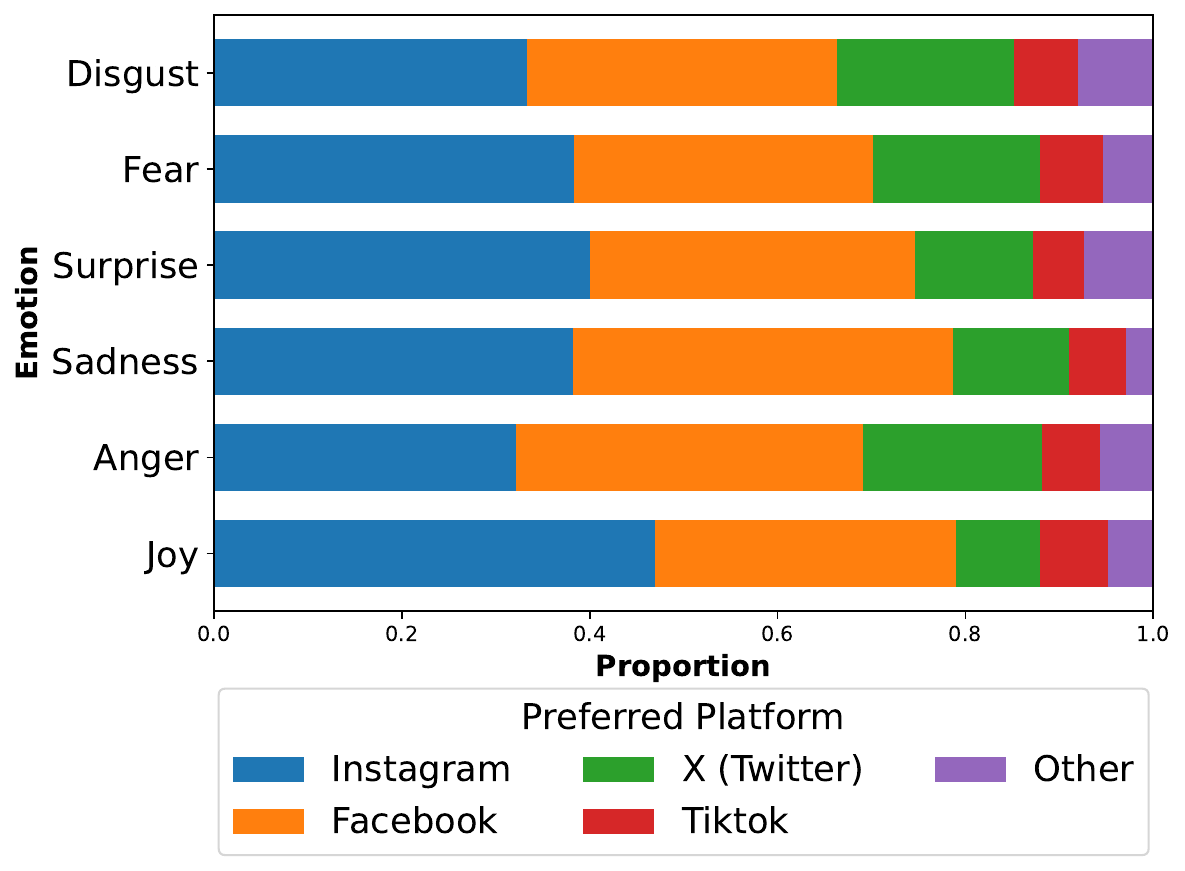}
	\caption{Authors' reported preferred social media platforms by emotion.}
	\label{preferred_platform_proportions}
\end{figure}

\subsection{Author Intent}
Authors'reasons for posting and social media habits vary by primary emotion too. While intended post audience varies little by emotion, see Figure~\ref{post_audience_counts_fig}, authors' purpose for posting varies greatly by emotion. We show this in Figure~\ref{why_post_counts} which illustrates the distribution of reported reasons for posting across different emotions. \emotion{Joy} posts are more likely to be posted to share an experience from their personal lives, while \emotion{anger}, \emotion{disgust}, and \emotion{fear} are more likely to advocate for or attack a subject or express controversial views. This aligns with the above findings that \emotion{joy} and \emotion{surprise} posts are more likely to use personal photos and have text that directly describes the image, while \emotion{anger}, \emotion{disgust}, and \emotion{fear} posts are more likely to use memes, professional photos, or screenshots.

Authors' behavior and preferences on social media also vary by emotion. While authors of different primary emotion posts report little difference in how often  they view social media, see Figure~\ref{social_media_use_proportions}, their posting frequency does differ. This is shown in Figure~\ref{social_media_post_freq_proportions}; authors of \emotion{joy} posts report posting less frequently than authors of all other emotions, with 40\% of \emotion{joy} authors posting less than once per week compared to 25-30\% of authors of other emotions. Furthermore, authors' preferred social media platform varies by emotion, as seen in Figure~\ref{preferred_platform_proportions}. Facebook and Instagram are the two most popular platforms for all emotions, however, Twitter(X) is twice as popular for authors of \emotion{anger}, \emotion{disgust}, and \emotion{fear} posts than for authors of \emotion{joy} posts.

\subsection{Discussion}

As we alluded to in our results, several patterns emerge across our experiments between emotions and how readers and models perform when reconstructing them. Here we discuss the relationships between emotions that we observe in our experiments and how they relate to theories of emotion.

\emotion{Joy} and \emotion{surprise} have an asymmetric relationship: \emotion{joy} is nearly the easiest emotion for readers and models to reconstruct, while \emotion{surprise} is the most difficult. \emotion{Joy} is unlikely to have \emotion{surprise} as a secondary emotion (17\% of \emotion{joy} posts express surprise), while \emotion{surprise} is likely to have joy as a secondary emotion (62\% of \emotion{surprise} posts express joy). Readers struggle with \emotion{joy} posts when more than one emotion is expressed (45 F1 point drop) while performance is slightly higher for \emotion{surprise} posts with more than one emotion (2 F1 point increase). \emotion{Joy} posts are the least likely to have a different emotion as the primary experienced emotion but have the highest difference in performance when the mismatch occurs, while \emotion{surprise} posts are the most likely to have a different emotion as the primary experienced emotion but have the lowest difference in performance when the mismatch occurs. 

We attribute this to theoretical differences in the nature of these emotions. Joy is a sustained, valence-stable, goal-congruent emotion \cite{Ellsworth_Scherer_2003}, while surprise is a brief transitional or interrupt emotion \cite{ekman1999basic}. Joy is a highly prototypical emotion, with a clear set of cultural markers \cite{Fehr_Russell_1984}, which we believe leads to the strong performance we see from readers and models. However this prototypicality is a double-edged sword, as when other emotions are present or when the primary experienced emotion is not joy, reconstruction becomes more difficult. Surprise, on the other hand, is an ambiguous state which resolves into other emotions, which explains why readers and models perform better when other emotions are present; because surprise is inherently underspecified, additional emotion cues in the post may actually constrain the interpretation.

\emotion{Anger} and \emotion{Disgust} also have a relationship, as they are often expressed together and have similar performance patterns. Both are present in about 50\% of the other emotion's posts, are similarly difficult for readers to reconstruct, have similar numbers of mismatches with experienced emotions, are of similar difficulty for readers to understand the triggering event and have similar patterns of performance based on the presence of multiple emotions. The content of the posts are also similar, as they have similar distributions of image types and relationships between text and image, and have nearly identical counts of secondary emotions expressed in their posts. 

Unlike the joy-surprise relationship, anger and disgust share a fundamentally similar appraisal profile. Both are elicited primarily by perceived norm violations: unfairness, offensive behavior, and moral transgressions \cite{rozin1999cad}. The CAD Triad groups contempt, anger, and disgust as the three canonical moral emotions, each linked to violations of different moral codes (autonomy, divinity, community) \cite{rozin1999cad}. Anger and disgust sit closest together within this triad in terms of the stimuli that trigger them.

Because they are triggered by overlapping classes of events, posts expressing one will very often also warrant the other, which explains the ~50\% mutual co-occurrence. A post about political corruption, cruelty, or injustice naturally evokes both: anger at the perpetrator and disgust at the act itself.

We suggest every observed pattern, co-occurrence, reconstruction difficulty, triggering event ambiguity, content similarity, follows from this shared moral appraisal structure. The difficulty of reconstruction is not about reader or model failure per se, but about a genuine underdetermination in the signal: posts often do not carry enough information to reliably discriminate between the two. This is supported by the fact that \emotion{disgust} and \emotion{anger} show the largest improvement in reader performance when the reader's event description is similar to the author's event description, as this additional information can help to disambiguate between the two emotions.

\section{Additional Modeling Details}
\subsection{Model Details}
\label{app:modeldetails}

All models are trained using four NVIDIA L40 GPUs. Each supervised model is fine-tuned three times, using the exact same setup and
environment.  The average and range of scores is reported.

We fine-tune the clip-vit-base-patch32 model using a combination of transformers and torchvision Python packages.\footnote{\url{https://huggingface.co/openai/clip-vit-base-patch32}}
Both images and text are encoded via the CLIP processor, and then fused using simple concatenation. All text+vision models are trained using 5 epochs,
1e-5 learning rate, batch size 8, cross-entropy loss, and early stopping. 

Appraisal models are trained using a multihead regression head (21, one for each appraisal dimension) on top of the same CLIP-based architecture, using mean squared error loss. All other training settings are the same as above.

\subsection{Modeling Results}
\label{sec:model_eval_appendix}

\subsubsection{RQ1: Appraisals}
\label{sec:appraisal_appendix}
Table~\ref{table:appraisal_simple} shows the full prediction results for all appraisal dimensions for both, human readers and models.  Table~\ref{table:appraisal_regression_event_description} contains the full modeling results for appraisals using event descriptions as input.

\subsubsection{RQ2: Event Description Similarity}
\label{Appendix:event_sim}
Table~\ref{table:post_event_similarity_bins_full} shows the full results for our experiment breaking down reader performance by event understanding.

\begin{table}[t]
	\small 
\centering
\begin{tabular}{lrrrr}
\toprule
 & \textbf{Q1} & \textbf{Q2} & \textbf{Q3} & \textbf{Q4} \\
\midrule
Mean & .02 & .21 & .39 & .59 \\
Std. & .02 & .06 & .05 & .08 \\
Min. & .01 & .10 & .30 & .48 \\
Max & .10 & .30 & .48 & .97 \\
\midrule
Post Emo. (F1)&&&&\\

Anger & .36 & .42 & .45 & .56 \\
Disgust & .23 & .38 & .37 & .51 \\
Fear & .36 & .38 & .54 & .63 \\
Joy & .41 & .56 & .65 & .71 \\
Sadness & .41 & .54 & .65 & .79 \\
Surprise & .27 & .38 & .34 & .42 \\
Overall & .34 & .44 & .51 & .61 \\
\midrule
Appr. (RMSE) &&&&\\
Pleas. & 1.72 & 1.38 & 1.14 & 0.99 \\
Unpleas. & 1.82 & 1.53 & 1.35 & 1.18 \\
Not consid. & 1.80 & 1.61 & 1.58 & 1.45 \\
Own resp. & 1.70 & 1.53 & 1.44 & 1.29 \\
Other resp. & 1.92 & 1.75 & 1.80 & 1.86 \\
Chance ctrl. & 1.91 & 1.95 & 1.91 & 1.89 \\
Sit. resp. & 1.87 & 1.77 & 1.80 & 1.81 \\
Others ctrl. & 1.87 & 1.75 & 1.76 & 1.83 \\
Own ctrl. & 1.67 & 1.48 & 1.45 & 1.39 \\
Antici. cons. & 1.73 & 1.68 & 1.72 & 1.73 \\
Goal relev. & 1.82 & 1.69 & 1.69 & 1.67 \\
Goal support & 1.68 & 1.51 & 1.46 & 1.35 \\
Attention & 1.76 & 1.72 & 1.69 & 1.68 \\
Event pred. & 1.71 & 1.71 & 1.67 & 1.56 \\
External stnd. & 1.74 & 1.59 & 1.54 & 1.44 \\
Internal stnd. & 1.78 & 1.62 & 1.44 & 1.46 \\
Familiarity & 1.90 & 1.99 & 1.87 & 1.83 \\
Suddenness & 1.84 & 1.78 & 1.81 & 1.82 \\
Urgency & 1.96 & 1.93 & 1.93 & 1.92 \\
Effort & 1.83 & 1.67 & 1.68 & 1.58 \\
Overall & 1.81 & 1.70 & 1.65 & 1.60 \\
\bottomrule
\end{tabular}
\caption{Reader F1 and RMSE scores split by quartiles of author and reader event description similarity. Appraisal results only include dimensions with at least a 0.1 difference in RMSE between Q1 and Q4. }
\label{table:post_event_similarity_bins_full}
\end{table}

{\setlength{\tabcolsep}{5pt}
\begin{table*}[t]
	\small 
	\centering
\begin{tabular}{lccccccccccccc}
    \toprule
& \multicolumn{2}{c}{Reader $\downarrow$} & \multicolumn{3}{c}{CLIP $\downarrow$} & \multicolumn{3}{c}{Qwen $\downarrow$} & \multicolumn{3}{c}{Random Image $\downarrow$} & \multicolumn{2}{c}{Baselines $\downarrow$} \\
\cmidrule(lr){2-3}\cmidrule(lr){4-6}\cmidrule(lr){7-9}\cmidrule(lr){10-12}\cmidrule(lr){13-14}
& Mean & Individual & T & I & T+I & T & I & T+I & T & I & T+I & Random & Mean \\
\cmidrule(lr){2-3}\cmidrule(lr){4-6}\cmidrule(lr){7-9}\cmidrule(lr){10-12}\cmidrule(lr){13-14}

    Accept conseq. & 1.63 & 1.95 & 1.44 & 1.48 & 1.47 & 1.81 & 1.98 & 1.88 & 1.44 & 1.42 & 1.44 & 2.05 & 1.41 \\
    Anticipated conseq. & 1.45 & 1.72 & 1.33 & 1.39 & 1.34 & 1.68 & 1.76 & 1.78 & 1.32 & 1.32 & 1.33 & 2.02 & 1.31 \\
    Attention & 1.44 & 1.72 & 1.38 & 1.41 & 1.40 & 1.81 & 1.78 & 1.89 & 1.38 & 1.36 & 1.38 & 1.96 & 1.35 \\
    Chance control & 1.63 & 1.92 & 1.48 & 1.51 & 1.49 & 1.80 & 1.79 & 1.99 & 1.48 & 1.46 & 1.48 & 2.15 & 1.46 \\
    Effort & 1.45 & 1.69 & 1.39 & 1.48 & 1.41 & 1.69 & 1.72 & 1.67 & 1.40 & 1.42 & 1.39 & 2.03 & 1.42 \\
    Event predic. & 1.42 & 1.66 & 1.27 & 1.33 & 1.30 & 1.55 & 1.74 & 1.59 & 1.27 & 1.29 & 1.29 & 2.01 & 1.28 \\
    External stan. & 1.42 & 1.58 & 1.38 & 1.47 & 1.40 & 1.63 & 1.65 & 1.63 & 1.38 & 1.49 & 1.40 & 2.21 & 1.48 \\
    Familiarity & 1.62 & 1.90 & 1.35 & 1.42 & 1.38 & 2.05 & 2.33 & 2.20 & 1.34 & 1.37 & 1.35 & 2.03 & 1.36 \\
    Goal relevance & 1.46 & 1.72 & 1.41 & 1.48 & 1.43 & 2.21 & 2.06 & 2.30 & 1.42 & 1.43 & 1.42 & 2.06 & 1.42 \\
    Goal support & 1.28 & 1.50 & 1.23 & 1.44 & 1.23 & 1.54 & 1.84 & 1.53 & 1.24 & 1.51 & 1.23 & 2.18 & 1.50 \\
    Internal stan. & 1.35 & 1.58 & 1.33 & 1.53 & 1.33 & 1.65 & 1.64 & 1.60 & 1.32 & 1.65 & 1.34 & 2.16 & 1.64 \\
    Not consider & 1.41 & 1.61 & 1.27 & 1.36 & 1.28 & 1.68 & 1.58 & 1.65 & 1.26 & 1.35 & 1.26 & 2.11 & 1.35 \\
    Other respon. & 1.62 & 1.84 & 1.60 & 1.70 & 1.61 & 1.83 & 1.88 & 1.88 & 1.59 & 1.63 & 1.59 & 2.13 & 1.62 \\
    Others control & 1.54 & 1.80 & 1.51 & 1.61 & 1.52 & 1.79 & 1.83 & 1.81 & 1.50 & 1.59 & 1.50 & 2.13 & 1.58 \\
    Own control & 1.28 & 1.50 & 1.22 & 1.31 & 1.23 & 1.50 & 1.76 & 1.53 & 1.22 & 1.31 & 1.22 & 2.14 & 1.31 \\
    Own respon. & 1.28 & 1.50 & 1.19 & 1.31 & 1.19 & 1.45 & 1.55 & 1.41 & 1.17 & 1.32 & 1.18 & 2.22 & 1.32 \\
    Pleasantness & 1.17 & 1.33 & 1.21 & 1.51 & 1.19 & 1.48 & 1.94 & 1.45 & 1.21 & 1.68 & 1.22 & 2.25 & 1.68 \\
    Situational respon. & 1.58 & 1.81 & 1.47 & 1.49 & 1.46 & 1.69 & 1.79 & 1.78 & 1.45 & 1.44 & 1.46 & 2.13 & 1.43 \\
    Suddenness & 1.59 & 1.81 & 1.50 & 1.57 & 1.51 & 1.64 & 1.71 & 1.63 & 1.49 & 1.51 & 1.48 & 2.09 & 1.50 \\
    Unpleasantness & 1.29 & 1.49 & 1.29 & 1.54 & 1.27 & 1.59 & 1.79 & 1.54 & 1.28 & 1.62 & 1.28 & 2.11 & 1.62 \\
    Urgency & 1.67 & 1.93 & 1.55 & 1.58 & 1.59 & 1.71 & 1.77 & 1.74 & 1.56 & 1.49 & 1.55 & 2.10 & 1.49 \\
\cmidrule(lr){2-3}\cmidrule(lr){4-6}\cmidrule(lr){7-9}\cmidrule(lr){10-12}\cmidrule(lr){13-14}
    Overall & 1.46 & 1.70 & 1.38 & 1.48 & 1.39 & 1.71 & 1.81 & 1.75 & 1.37 & 1.47 & 1.38 & 2.11 & 1.46 \\
\bottomrule
\end{tabular}
	\caption{RMSE scores of appraisal dimensions. Readers are evaluated using two approaches: mean uses the mean of the three readers and individual compares every reader separately to the authors. For CLIP models the mean RMSE of three runs is reported for each modality: text (T), image (I), multimodal (M). Baselines include a random baseline which applies random scores between 1 and 5 for each appraisal and a mean baseline which uses the mean score for each appraisal in the dataset for every instance.}
	\label{table:appraisal_simple}
\end{table*}
}

\begin{table*}[t]
	\centering
	\small
	\begin{tabular}{lcccccccc}
		\toprule
		& \multicolumn{3}{c}{Post Text} & \multicolumn{3}{c}{Event Description} & \multicolumn{2}{c}{Baselines} \\
		& T & I & T+I & T & I & T+I & Random & Mean \\
		\cmidrule(lr){2-4}\cmidrule(lr){5-7}\cmidrule(lr){8-9}
		Accept consequences & 1.42 & 1.44 & 1.42 & 1.44 & 1.47 & 1.46 & 2.08 & 1.41 \\
		Anticipated consequences & 1.30 & 1.33 & 1.32 & 1.29 & 1.37 & 1.33 & 2.04 & 1.31 \\
		Attention & 1.35 & 1.35 & 1.37 & 1.33 & 1.40 & 1.34 & 1.96 & 1.35 \\
		Chance control & 1.44 & 1.46 & 1.46 & 1.44 & 1.49 & 1.46 & 2.16 & 1.46 \\
		Effort & 1.36 & 1.41 & 1.37 & 1.33 & 1.47 & 1.37 & 2.03 & 1.42 \\
		Event predictability & 1.25 & 1.28 & 1.28 & 1.21 & 1.33 & 1.24 & 1.99 & 1.28 \\
		External standards & 1.37 & 1.45 & 1.37 & 1.32 & 1.46 & 1.34 & 2.17 & 1.48 \\
		Familiarity & 1.32 & 1.37 & 1.35 & 1.32 & 1.43 & 1.34 & 2.04 & 1.36 \\
		Goal relevance & 1.38 & 1.43 & 1.39 & 1.35 & 1.48 & 1.37 & 2.06 & 1.42 \\
		Goal support & 1.21 & 1.40 & 1.23 & 1.15 & 1.41 & 1.18 & 2.19 & 1.50 \\
		Internal standards & 1.29 & 1.48 & 1.28 & 1.24 & 1.50 & 1.24 & 2.18 & 1.64 \\
		Not consider & 1.24 & 1.31 & 1.24 & 1.22 & 1.35 & 1.23 & 2.09 & 1.35 \\
		Other responsibility & 1.56 & 1.63 & 1.55 & 1.53 & 1.71 & 1.55 & 2.19 & 1.62 \\
		Others control & 1.48 & 1.53 & 1.46 & 1.40 & 1.60 & 1.42 & 2.14 & 1.58 \\
		Own control & 1.20 & 1.27 & 1.21 & 1.14 & 1.29 & 1.16 & 2.14 & 1.31 \\
		Own responsibility & 1.16 & 1.28 & 1.17 & 1.13 & 1.29 & 1.14 & 2.24 & 1.32 \\
		Pleasantness & 1.19 & 1.47 & 1.19 & 1.03 & 1.47 & 1.03 & 2.28 & 1.68 \\
		Situational responsibility & 1.42 & 1.44 & 1.44 & 1.40 & 1.47 & 1.42 & 2.14 & 1.43 \\
		Suddenness & 1.45 & 1.51 & 1.48 & 1.45 & 1.57 & 1.49 & 2.08 & 1.50 \\
		Unpleasantness & 1.26 & 1.48 & 1.26 & 1.17 & 1.50 & 1.18 & 2.12 & 1.62 \\
		Urgency & 1.52 & 1.51 & 1.55 & 1.56 & 1.58 & 1.58 & 2.08 & 1.49 \\
		\cmidrule(lr){2-4}\cmidrule(lr){5-7}\cmidrule(lr){8-9}
		Overall & 1.35 & 1.42 & 1.36 & 1.31 & 1.46 & 1.34 & 2.12 & 1.46 \\
		\bottomrule
	\end{tabular}
	\caption{Model performance (RMSE) for appraisal regression on Mult2Emo test set using event descriptions as input. Models trained either predict all appraisals or jointly predict all appraisals and emotion labels. Models are fine-tuned on text only (T), image only (I), and text + image (T+I).}
	\label{table:appraisal_regression_event_description}
\end{table*}

\subsection{MLLM Prompts}
The prompts used in our experiments with multimodal large language models (MLLMs) are provided in Table~\ref{MLLM_Prompts}. 

\begin{table*}
  \centering\sffamily\small\scalefont{0.85}
  \begin{tabularx}{\linewidth}{@{}c p{10mm} XXX}
    \toprule
    & Section & Text & Image & Text + Image \\
    \cmidrule(r){2-2}\cmidrule(rl){3-3}\cmidrule(rl){4-4}\cmidrule(rl){5-5}
    \multirow{4}{*}{\rotatebox[origin=c]{90}{Emotion}}
            &Task \newline Descr.
            &Which of the following emotions is the author of the post trying to express?
                   &Which of the following emotions is the author of the post trying to express?
                            &Which of the following emotions is the author of the post trying to express?
    \\    \cmidrule(r){2-2}\cmidrule(rl){3-3}\cmidrule(rl){4-4}\cmidrule(rl){5-5}
            &Labels
            &\{Emotions\}
    "Answer with exactly one word: the single emotion from the list above.
                   &\{Emotions\}
    "Answer with exactly one word: the single emotion from the list above.
                            &\{Emotions\}
    "Answer with exactly one word: the single emotion from the list above.
    \\    \cmidrule(r){2-2}\cmidrule(rl){3-3}\cmidrule(rl){4-4}\cmidrule(rl){5-5}
            &Format \newline Instr.
            &Do not explain, do not add punctuation, do not write anything else.
                   &Do not explain, do not add punctuation, do not write anything else.
                            &Do not explain, do not add punctuation, do not write anything else.
    \\    \cmidrule(r){2-2}\cmidrule(rl){3-3}\cmidrule(rl){4-4}\cmidrule(rl){5-5}
            &Data \newline Input
            &Below is the text of a social media post. \{text\}
                   & Above is the image from a social media post.
                            &Above is the image from a social media post, and below is its text.\\

    \midrule

    \multirow{5}{*}{\rotatebox[origin=c]{90}{Appraisals}}
            &Task \newline Descr.
            &Consider the event that the post is about, and how the author of the post
    experienced it. Rate each statement below on a scale from 1 to 5
                   &Consider the event that the post is about, and how the author of the post
    experienced it. Rate each statement below on a scale from 1 to 5
                            &Consider the event that the post is about, and how the author of the post
    experienced it. Rate each statement below on a scale from 1 to 5
    \\    \cmidrule(r){2-2}\cmidrule(rl){3-3}\cmidrule(rl){4-4}\cmidrule(rl){5-5}
            &Labels
            &1 = Not at all, 2 = A little, 3 = Moderately, 4 = Quite a bit, 5 = Extremely
                   &1 = Not at all, 2 = A little, 3 = Moderately, 4 = Quite a bit, 5 = Extremely
                            &1 = Not at all, 2 = A little, 3 = Moderately, 4 = Quite a bit, 5 = Extremely
    \\    \cmidrule(r){2-2}\cmidrule(rl){3-3}\cmidrule(rl){4-4}\cmidrule(rl){5-5}
            &Appraisal \newline Stat.
            & \{Appraisal Statements\}
                   & \{Appraisal Statements\}
                            & \{Appraisal Statements\} \\
                              \cmidrule(r){2-2}\cmidrule(rl){3-3}\cmidrule(rl){4-4}\cmidrule(rl){5-5}
            &Format \newline Instr.
            &Answer with exactly {n} lines, one per statement, in the format "<statement number>. <rating>" (for example "1.~3").
    Give a rating for every statement.
    Do not explain, do not repeat the statements, do not write anything else.
                   &Answer with exactly {n} lines, one per statement, in the format "<statement number>. <rating>" (for example "1.~3").
    Give a rating for every statement.
    Do not explain, do not repeat the statements, do not write anything else.
                            &Answer with exactly {n} lines, one per statement, in the format "<statement number>. <rating>" (for example "1.~3").
    Give a rating for every statement.
    Do not explain, do not repeat the statements, do not write anything else.
    \\    \cmidrule(r){2-2}\cmidrule(rl){3-3}\cmidrule(rl){4-4}\cmidrule(rl){5-5}
            &Data \newline Input
            &Below is the text of a social media post. \{text\}
                   & Above is the image from a social media post.
                            &Above is the image from a social media post, and below is its text.\\
    \bottomrule
  \end{tabularx}
  \caption{Prompts for text, image, and text + image modalities, for both emotion classification and appraisal prediction. Variables are typeset in \{curly
    brackets\}. Emotion labels and appraisal statements can be found in Table~\ref{tab:survey_options}. The ordering of each is randomized per prompting instance.}
  \label{MLLM_Prompts}
\end{table*}

\end{document}